\documentclass{article}

\usepackage{CJKutf8}

\usepackage{xspace}

\usepackage{amsmath,amssymb,amsfonts}
\usepackage{algorithmic}
\usepackage{graphicx}
\usepackage{textcomp}
\usepackage{xcolor}
\usepackage{algorithm}
\usepackage{wrapfig}
\usepackage{booktabs}
\usepackage{lipsum}
\usepackage{balance}
\usepackage{url}
\usepackage{chngpage}

\usepackage{enumitem}
\usepackage{verbatimbox}
\usepackage{tcolorbox}
\usepackage{fancyvrb}
\usepackage{fancyref}
\usepackage{caption}
\usepackage{subcaption}
\usepackage[subtle]{savetrees}
\usepackage{soul}

\usepackage{inconsolata}
\usepackage[T1]{fontenc}
\usepackage{array}
\usepackage{pifont}
\usepackage{xcolor}

\newcommand{\eg}{e.g., }

\newcommand{\greentick}{\textcolor{state-green}{\ding{51}}}
\newcommand{\redcross}{\textcolor{red}{\ding{55}}}

\newcolumntype{P}[1]{>{\centering\arraybackslash}p{#1}}
\newcolumntype{M}[1]{>{\centering\arraybackslash}m{#1}}
\newcolumntype{B}[1]{>{\centering\arraybackslash}b{#1}}

\definecolor{light-gray}{rgb}{0.8, 0.8, 0.8}
\definecolor{comment-green}{rgb}{0.435, 0.576, 0.106}
\definecolor{state-green}{rgb}{0.435, 0.576, 0.106}
\definecolor{prompt-blue}{HTML}{2596be}
\definecolor{execute-code-red}{HTML}{f4cccc}
\definecolor{execute-lm-purple}{HTML}{d9d2e9}
\definecolor{code-function}{HTML}{693da8}  
\definecolor{code-syntax}{HTML}{0060b1}
\definecolor{code-constant}{HTML}{d86001}
\definecolor{prompt-gray}{HTML}{a7a7a7}
\definecolor{highlight}{HTML}{f8f9cb}
\definecolor{highlight}{HTML}{e3eeff}  
\definecolor{code-perception}{HTML}{2ecc71}
\definecolor{code-control}{HTML}{ff9900}
\definecolor{code-undefined}{HTML}{ff0000}
\renewcommand\fbox{\fcolorbox{light-gray}{white}}

\newcommand{\algfullname}{Chain of Code\xspace}
\newcommand{\algname}{CoC\xspace}

\newcommand{\hlgen}[1]{\colorbox{highlight}{\parbox{0.95\linewidth}{#1}}}
\newcommand{\hlruncode}[1]{\colorbox{execute-code-red}{\parbox{0.95\linewidth}{#1}}}
\newcommand{\hlrunlm}[1]{\colorbox{execute-lm-purple}{\parbox{0.95\linewidth}{#1}}}

\newcommand{\linenumber}[1]{\textcolor{gray}{#1}\ \ \ \ }

\usepackage{microtype}
\usepackage{graphicx}
\usepackage{booktabs} 

\usepackage{hyperref}

\usepackage{stfloats}

\usepackage[accepted]{icml2024}

\usepackage{amsmath}
\usepackage{amssymb}
\usepackage{mathtools}
\usepackage{amsthm}

\usepackage[capitalize,noabbrev]{cleveref}

\theoremstyle{plain}

\theoremstyle{definition}

\theoremstyle{remark}

\usepackage[textsize=tiny]{todonotes}

\icmltitlerunning{\algfullname: Reasoning with a Language Model-Augmented Code Emulator}

\begin{document}
\begin{CJK*}{UTF8}{gbsn}

\twocolumn[
\icmltitle{\algfullname: Reasoning with a Language Model-Augmented Code Emulator}



\icmlsetsymbol{equal}{*}

\begin{icmlauthorlist}
\icmlauthor{Chengshu Li}{stanford}
\icmlauthor{Jacky Liang}{gdm}
\icmlauthor{Andy Zeng}{gdm}
\icmlauthor{Xinyun Chen}{gdm}
\icmlauthor{Karol Hausman}{stanford,gdm}
\icmlauthor{Dorsa Sadigh}{stanford,gdm}
\icmlauthor{Sergey Levine}{gdm,berkeley}
\icmlauthor{Li Fei-Fei}{stanford}
\icmlauthor{Fei Xia}{equal,gdm}
\icmlauthor{Brian Ichter}{equal,gdm}
\end{icmlauthorlist}

\centering
\url{https://chain-of-code.github.io/}

\icmlaffiliation{stanford}{Department of Computer Science, Stanford University, California, USA}
\icmlaffiliation{gdm}{Google DeepMind, California, USA}
\icmlaffiliation{berkeley}{Department of Electrical Engineering and Computer Sciences, University of California, Berkeley, California, USA}

\icmlcorrespondingauthor{Chengshu Li}{chengshu@stanford.edu}

\icmlkeywords{Machine Learning, ICML}

\vskip 0.3in
]



\printAffiliationsAndNotice{\icmlEqualContribution} 

\begin{abstract}
%
Code provides a general syntactic structure to build complex programs and perform precise computations when paired with a code interpreter - we hypothesize that language models (LMs) can leverage code-writing to improve Chain of Thought reasoning not only for logic and arithmetic tasks \cite{chen2022program,nye2021show,austin2021program}, but also for \textit{semantic} ones (and in particular, those that are a mix of both).
For example, consider prompting an LM to write code that counts the number of times it detects sarcasm in an essay: the LM may struggle to write an implementation for ``\texttt{detect\_sarcasm(string)}'' that can be executed by the interpreter (handling the edge cases would be insurmountable). However, LMs may still produce a valid solution if they not only write code, but also selectively ``emulate'' the interpreter by generating the expected output of ``\texttt{detect\_sarcasm(string)}''. 
In this work, we propose \algfullname (\algname), a simple yet surprisingly effective extension that improves LM code-driven reasoning.  The key idea is to encourage LMs to format semantic sub-tasks in a program as flexible pseudocode that the interpreter can explicitly catch undefined behaviors and hand off to simulate with an LM  (as an ``LMulator").
Experiments demonstrate that \algfullname outperforms Chain of Thought and other baselines across a variety of benchmarks; on BIG-Bench Hard, \algfullname achieves 84\%, a gain of 12\% over Chain of Thought. In a nutshell, \algname broadens the scope of reasoning questions that LMs can answer by ``thinking in code".
\vspace{-5mm}

\end{abstract}

\begin{figure}[t]
\centering
\begin{subfigure}{\linewidth}
\caption*{Direct answer only}
  \setlength{\fboxsep}{3pt}
  \noindent\fbox{\parbox{\linewidth}{\scriptsize{\texttt{{
  \\
  {Q: How many countries have I been to? I've been to Mumbai, London, Washington, Grand Canyon, ...}\\ 
  \hlgen{\textbf{A: 32 (20\%, \redcross), 29 (10\%, \redcross), 52 (10\%, \greentick), ...}}}\\
  }}}}\\
\caption*{Chain of Thought}
  \setlength{\fboxsep}{3pt}
  \noindent\fbox{\parbox{\linewidth}{\scriptsize{\texttt{{
  \\
  {Q: Let's think step by step. How many countries have I been to? I've been to Mumbai, London, ...}\\ 
  \hlgen{We'll group by countries and count:\\
1. India: Mumbai, Delhi, Agra\\
2. UK: London, Dover, Edinburgh, Skye\\
3. USA: Washington, Grand Canyon, ...\\
\textbf{A: 61 (20\%, \redcross), 60 (20\%, \redcross), 52 (10\%, \greentick), ...}}\\
  }}}}}\\
%
\caption*{\algfullname (Ours)}
  \setlength{\fboxsep}{3pt}
  \noindent\fbox{\parbox{\linewidth}{\scriptsize{\texttt{{
  \\
  {Q: How many countries have I been to? I've been to Mumbai, London, Washington, Grand Canyon, Baltimore, ...}\\ 
  \hlruncode{\linenumber{1}places, countries = ["Mumbai", ...], set() \\{\color{state-green}delta state: \{places = [`Mumbai', ...], countries = set()\}}}\\
  \hlruncode{\textcolor{gray}{2}\ \ \ \ for place in places: \\{\color{state-green}delta state: \{place = `Mumbai'\}}}\\
  \hlrunlm{\textcolor{gray}{3}\ \ \ \ \ \ country = get\_country(place) \\{\color{state-green}delta state: \{country = `India')\}}}\\
  \hlruncode{\textcolor{gray}{4}\ \ \ \ \ \ countries.add(country) \\{\color{state-green}delta state: \{countries = \{`India'\}\}}}\\
  \hlruncode{\textcolor{gray}{5}\ \ \ \ answer = len(countries) \ \ \ {\color{state-green}delta state: \{answer = 52\}}}\\
  \textbf{{A: 52 (100\%, \greentick)}}
  }}}}}\\
\end{subfigure}
     \vspace{-6mm}
     \caption{
     \algfullname generates code and reasons through an LM-augmented code emulator. Lines evaluated with Python are in \colorbox{execute-code-red}{red} and with an LM are in \colorbox{execute-lm-purple}{purple}. The full query is in Fig.~\ref{fig:intro_query}.
     }
    \vspace{-5mm}
\label{fig:intro}
\end{figure}

\begin{figure*}[t]
\centering

    \begin{subfigure}[b]{0.32\textwidth}
         \captionsetup{singlelinecheck=off, justification=raggedright, margin={50pt, 0pt}}
         \centering
         \includegraphics[width=\textwidth]{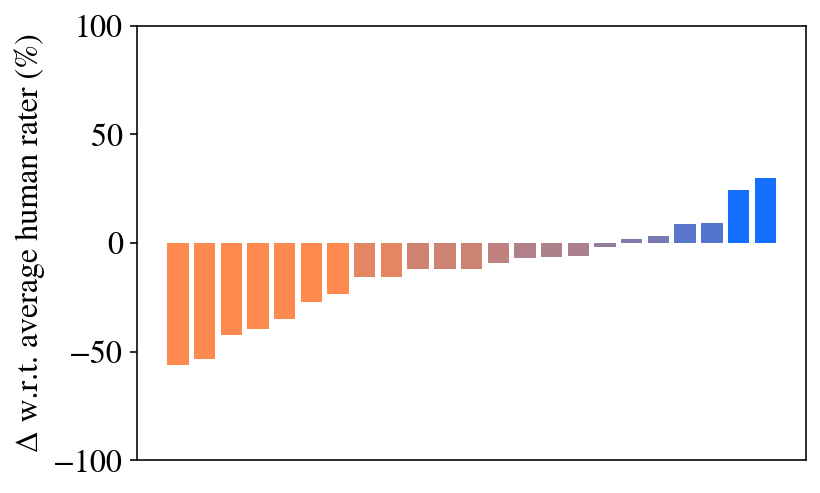}
         \vspace{-7mm}
         \caption{Direct answer only}
         \label{fig:all_tasks_direct}
     \end{subfigure}
     \hfill
     \begin{subfigure}[b]{0.32\textwidth}
         \captionsetup{singlelinecheck=off, justification=raggedright, margin={50pt, 0pt}}
         \centering
         \includegraphics[width=\textwidth]{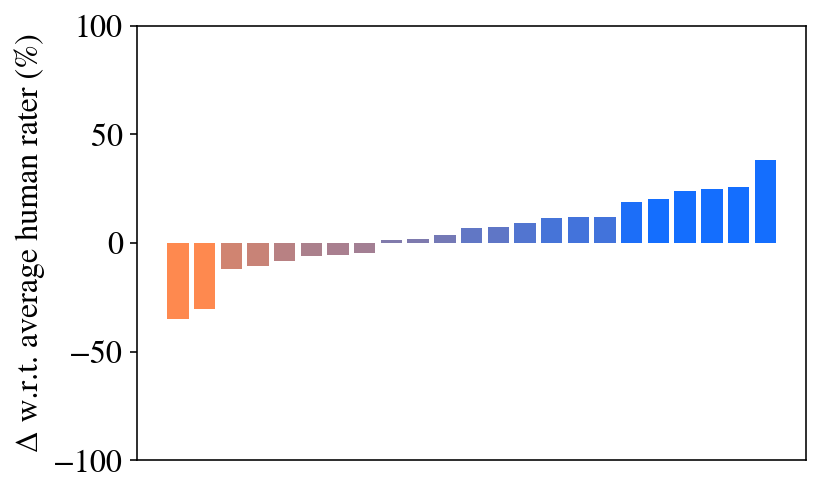}
         \vspace{-7mm}
         \caption{Chain of Thought}
         \label{fig:all_tasks_cot}
     \end{subfigure}
     \hfill
     \begin{subfigure}[b]{0.32\textwidth}
         \captionsetup{singlelinecheck=off, justification=raggedright, margin={40pt, 0pt}}
         \centering
         \includegraphics[width=\textwidth]{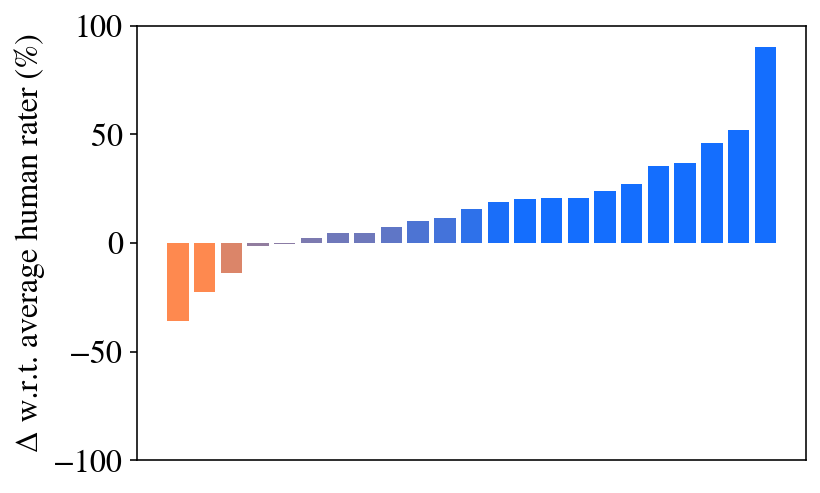}
         \vspace{-7mm}
         \caption{\algfullname (Ours)}
         \label{fig:all_tasks_coc_interweave_no_title}
     \end{subfigure}
     \vspace{-3mm}
     \caption{
     Overall results on BIG-Bench Hard compared to human performance~\citep{srivastava2022beyond}. 
     }
    \vspace{-3mm}
\label{fig:overall}
\end{figure*}

\section{Introduction}
\label{sec:intro}

Language models (LMs) at certain scale exhibit the profound ability to solve complex reasoning questions~\citep{brown2020language,wei2022emergent} – from writing math programs \cite{drori2022neural} to solving science problems \cite{lewkowycz2022solving}.
Notably, these capabilities have shown to improve with Chain of Thought (CoT) prompting~\citep{wei2022chain}, whereby complex problems are decomposed into a sequence of intermediate reasoning steps.
%
CoT excels at semantic reasoning tasks, but tends to struggle with questions that involve numeric or symbolic reasoning \cite{suzgun2022challenging,mirchandani2023large}.
Subsequent work addresses this by prompting LMs (\eg trained on Github~\citep{chen2021evaluating}) to write and execute code~\cite{chen2022program,nye2021show,austin2021program}.
%
Code in particular is advantageous because it provides both (i) a general syntactic structure to build and encode complex programs \cite{liang2023code} (\eg logic structures, functional vocabularies – in ways that are Turing complete), and (ii) an interface by which existing APIs paired together with an interpreter can be used to perform precise algorithmic computations (\eg from multiplication of large numbers to sorting an array of size 10,000) that a language model trained only to mimic the statistically most likely next token would otherwise struggle to produce.

While writing and executing code may improve LM reasoning performance across a wide range of arithmetic tasks, this particular approach contends with the fact that many semantic tasks are rather difficult (and at times, nearly impossible) to express in code.
For example, it remains unclear how to write a function that returns a boolean when it detects sarcasm in a string \cite{suzgun2022challenging} (handling the edge cases would be insurmountable).
%
Perhaps fundamentally, using LMs to write programs in lieu of multi-step textual reasoning inherently assumes that the intermediate reasoning traces (expressed in lines of code) all need to be \textit{executable} by an interpreter. Is it possible to lift these restrictions to get the best of both reasoning in code and reasoning in language?
%

In this work, we propose \algfullname (\algname), a simple yet surprisingly effective extension to improve LM code-driven reasoning -- where the LM not only writes a program, but also selectively ``simulates'' the interpreter by generating the expected output of certain lines of code (that the interpreter could not execute). The key idea is to encourage LMs to format semantic sub-tasks in a program as flexible pseudocode that at runtime can be explicitly caught and handed off to emulate with an LM -- we term this an LMulator (a portmanteau of LM and emulator). For example, given the task ``\textit{in the above paragraph, count how many times the person was sarcastic},'' we can in-context prompt the LM to write a program that may call helper functions such as \texttt{is\_sarcastic(sentence)}, to which the LM makes a linguistic prediction and returns the result as a boolean output, that then gets processed with the rest of the program.
Specifically, we formulate LM reasoning as the following process (illustrated in Figure~\ref{fig:intro}): the LM writes code, the interpreter steps through to execute each line of code (in \colorbox{execute-code-red}{red}), or if it fails, simulates the result with the LM (in \colorbox{execute-lm-purple}{purple}) and updates the program state (in {\color{state-green}green}).
\algname inherits the benefits of both (i) writing executable code (where precise algorithmic compututations are left to an interpreter), and (ii) writing pseudocode for semantic problems, and generating their outputs (which can be thought of as a simple formatting change, to which LMs are robust \cite{min2022rethinking}) -- enabling the LM to ``think in code''.

Extensive experiments demonstrate that \algname is applicable to a wide variety of challenging numerical and semantic reasoning questions, and outperforms a number of popular baselines.
In particular, we find that it achieves high performance on BIG-Bench Hard tasks~\citep{suzgun2022challenging}, outperforming average human raters overall and outperforming even the best human raters on an algorithmic subset of tasks, and to the best of our knowledge setting a new state of the art.
We further show that \textit{both} code interpreter execution and language model execution simulation are necessary for this performance, and that the approach scales well with large and small models alike -- contrary to prompting techniques like Chain of Thought that only emerge at scale. We then demonstrate how \algfullname can serve as a general purpose reasoner via \textit{cross-task prompting} benchmark, which in contrast to prior work, uses prompts from different families of problems as context -- providing only the structure of the response (as opposed to the solution itself). Finally, we show \algname is complementary to more advanced instruction tuned chat models, robust against prompt variation, and applicable beyond language reasoning domain like robotics.
This work underscores how one may leverage the structure and computational power of code and the reasoning abilities of language models to enable a ``best of both worlds'' reasoner.

\section{\algfullname: Reasoning with an LMulator}
\label{sec:coc}
In this section, we describe \algfullname (\algname), an approach that leverages the ability of language models to code, to reason, and to leverage an LM-augmented code emulator (an LMulator) to simulate running code. 
We start with background in Section~\ref{sec:prelim}, then overview the method in Section~\ref{sec:coc_method}, its implementation in Section~\ref{sec:implementation}, and finally its capabilities in Section~\ref{sec:abilities}.


\subsection{Preliminaries}
\label{sec:prelim}

Briefly, we overview some background on LM reasoning.
Many of these reasoning techniques have been enabled by in-context learning~\citep{brown2020language}, which provides the model with a few demonstrative examples at inference time, rather than updating any weights with gradients.
These examples serve to provide context and format for the setting, enabling the model to emulate these examples while adapting to a new query.
This property has been instrumental in easily applying LMs to new tasks as it can be rapidly adapted and requires minimal data.

Through in-context learning, approaches have been developed to leverage human thought processes and use tools to improve performance of language models.
We outline three such approaches that provide the foundations for \algfullname.
Chain of Thought (CoT)~\citep{wei2022chain}, ScratchPad~\citep{nye2021show}, and Program of Thoughts~\citep{chen2022program} demonstrated the efficacy of breaking problems down into substeps.
For CoT these substeps are in natural language, mirroring one's thought process when stepping through a complicated problem.
ScratchPad, on the other hand, maintains a program state of intermediate steps when simulating the output of code -- resulting in an LM acting as a code interpreter. 
Program of Thoughts~\citep{chen2022program} focused on generating the code itself, which is then executed by a code interpreter to solve reasoning problems.
Each of these is visualized in Figure~\ref{fig:prelim}.

\begin{figure*}[htb]
\centering
\begin{subfigure}{.28\linewidth}
\caption{Chain of Thought\vspace{-5pt}}
  \setlength{\fboxsep}{3pt}
  \flushleft
  \noindent\fbox{\parbox{0.93\linewidth}{\scriptsize{\texttt{{
  \\
  {Q: Roger has 5 balls. He buys 2 more packs, each with 3. How many balls does he have now?}\\ \\
  \hlgen{Roger starts with 5 balls.}
  \hlgen{2 packs of 3 balls is 6.}
  \hlgen{5 + 6 = 11.}\\
  \\ {A: 11}
  }}}}}\\
  \label{fig:prelim-cot}
\end{subfigure}%
%
\begin{subfigure}{.28\linewidth}
\caption{Program of Thoughts\vspace{-5pt}}
  \setlength{\fboxsep}{3pt}
  \flushright
  \noindent\fbox{\parbox{0.93\linewidth}{\scriptsize{\texttt{{
  \\
  {Q: Roger has 5 balls. He buys 2 more packs, each with 3. How many balls does he have now?}\\ \\
  \hlruncode{num\_balls = 5}\\
  \hlruncode{num\_balls += 2 * 3}\\
  \hlruncode{answer = num\_balls}\\
  \\ {A: 11}
  }}}}}\\
  \label{fig:prelim-pot}
\end{subfigure}
\hspace{-10pt}
\begin{subfigure}{.43\linewidth}
\caption{ScratchPad\vspace{-5pt}}
  \setlength{\fboxsep}{3pt}
  \flushright
  \noindent\fbox{\parbox{0.93\linewidth}{\scriptsize{\texttt{{
  \\
  {Q: Roger has 5 balls. He buys 2 more packs, each with 3. How many balls does he have now?}\\ \\ \\
  \hlrunlm{num\_balls = 5 \quad{\color{state-green}state: \{num\_balls = 5\}}}\\
  \hlrunlm{num\_balls += 2 * 3 \quad{\color{state-green}state: \{num\_balls = 11\}}}\\
  \hlrunlm{answer = num\_balls \quad{\color{state-green}state: \{answer = 11\}}}\\
  \\ {A: 11}
  }}}}}\\
  \label{fig:prelim-scratchpad}
\end{subfigure}
\vspace{2mm}
\hrule
\begin{subfigure}{.4\linewidth}
\caption{\algfullname Generation (Ours)}
\vspace{-2mm}
  \setlength{\fboxsep}{3pt}
  \flushleft
  \noindent\fbox{\parbox{0.93\linewidth}{\scriptsize{\texttt{{
  \\
  {Q: I have an orange, a violin, two peaches, an apple, a pepper, and three plums. How many fruits do I have?}\\ \\
  \hlgen{\linenumber{1}objects = \{"orange": 1, "violin": 1, "peaches": 2, "apple": 1, "pepper": 1, "plum": 3\}}
  \hlgen{\linenumber{2}num\_fruits = 0 \\}\\
  \hlgen{\linenumber{3}for object in objects: \\}\\
  \hlgen{\linenumber{4}\quad object\_is\_fruit = is\_fruit(object) \\}\\
  \hlgen{\linenumber{5}\quad if object\_is\_fruit: \\}\\
  \hlgen{\linenumber{6}\quad \quad num\_fruits += objects[object] \\}\\
  \hlgen{\linenumber{7}answer = num\_fruits \\}\\ \\
  \\ 
  }}}}}\\
  \label{fig:method_generation}
\end{subfigure}%
\begin{subfigure}{.57\linewidth}
\caption{\algfullname Execution (Ours)}
\vspace{-2mm}
  \setlength{\fboxsep}{3pt}
  \flushright
  \noindent\fbox{\parbox{0.93\linewidth}{\scriptsize{\texttt{{
  \\
  {Q: I have an orange, a violin, two peaches, an apple, a pepper, and three plums. How many fruits do I have?}\\ \\
  \hlruncode{\linenumber{1}objects = \{"orange": 1, "violin": 1, "peaches": 2, "apple": 1, "pepper": 1, "plum": 3\}  \\{\color{state-green}delta state: \{objects = \{`orange': 1, `violin': 1, ...\}\} }}\\
  \hlruncode{\linenumber{2}num\_fruits = 0 \\{\color{state-green}delta state: \{num\_fruits = 0\}}}\\
  \hlruncode{\linenumber{3}for object in objects: \\{\color{state-green}delta state: \{object = `orange'\}  \# updated for each loop}}\\
  \hlrunlm{\linenumber{4}\quad object\_is\_fruit = is\_fruit(object) \\{\color{state-green}delta state: \{object\_is\_fruit = True\}}}\\
  \hlruncode{\linenumber{5}\quad if object\_is\_fruit: \\{\color{state-green}delta state: \{\}}}\\
  \hlruncode{\linenumber{6}\quad \quad num\_fruits += objects[object] \\{\color{state-green}delta state: \{num\_fruits = 1\}}}\\
  \hlruncode{\linenumber{7}answer = num\_fruits\\{\color{state-green}delta state: \{answer = 7\}}}\\ \\
  \\ {A: 7}
  }}}}}\\
  \label{fig:method_execution}
\end{subfigure}
\caption{
{\bf Previous reasoning methods:} To solve advanced problems, (\ref{fig:prelim-cot}) Chain of Thought prompting breaks the problem down into intermediate steps, (\ref{fig:prelim-pot}) Program of Thoughts prompting writes and executes code, and (\ref{fig:prelim-scratchpad}) ScratchPad prompting simulates running already written code by tracking intermediate steps through a program state.
{\bf Our reasoning method:} \algfullname first (\ref{fig:method_generation}) generates code or psuedocode to solve the question and then (\ref{fig:method_execution}) executes the code with a code interpreter if possible, and with an LMulator (language model emulating code) otherwise. \colorbox{highlight}{Blue} highlight indicates LM generation, \colorbox{execute-code-red}{red} highlight indicates LM generated code being executed, and \colorbox{execute-lm-purple}{purple} highlight indicates LMulator simulating the code via a program state in {\color{state-green}green}. 
}
\label{fig:prelim}
\vspace{-3mm}
\end{figure*}

\subsection{\algfullname}
\label{sec:coc_method}

Inspired by how a human may reason through a particularly complex problem with a mix of natural language, pseudocode, and runnable code or how a researcher may develop a new general algorithm through a code-based formalism then apply it to a problem,
\algfullname proceeds in two steps: 
(1) Generation, which, given the question to solve, an LM generates code to reason through the problem, and 
(2) Execution, which executes the code via a code interpreter when possible and via an LM when not.
See Section~\ref{sec:implementation} for more details on the specific implementation.

\textbf{\algfullname Generation}
Given a problem to solve, \algname generates reasoning substeps in the structure of code.
This code provides the framework of reasoning through the problem, and may be in the form of explicit code, pseudocode, or natural language.
Figure~\ref{fig:method_generation} walks through a potential generation to solve an object counting problem from BIG-Bench. 

\textbf{\algfullname Execution}
A core contribution of \algname is not just the generation of reasoning code, but the manner in which it is executed. 
Once the code is written, the code is attempted to be run by a code interpreter -- in this work we consider Python, but the approach is general to any interpreter. 
If the code is successfully executed, the program state is updated and the execution continues.
If the code is not executable or raises any exception, the language model instead is used to simulate the execution.
The program state is subsequently updated by the language model's outputs and the execution continues.
Herein, we refer to this as an LMulator, a portmanteau of LM and code emulator.
This relatively simple change enables a variety of new applications for code which mix semantics and numerics.
Figure~\ref{fig:method_execution} shows how the generated code is run, maintaining the program state and switching between the Python executor and the LMulator.

\subsection{\algfullname Implementation}
\label{sec:implementation}
While the generation implementation is straightforward prompting and language model generation, the execution implementation is slightly more complex.
Our implementation is based on using Python's \texttt{try} and \texttt{except} and maintaining a program state.
Line by line \algname steps through the code.
If the line is executable by a code interpreter, it is executed, the program state is updated, and the program continues.
If it is not executable by a code interpreter, a language model is given the context of the program (the question, the prior lines, and the history of the program state) and generates the next program state.
This emulation can also leverage chain of thought to determine how to respond.
That generated program state is then updated for the code interpreter as well.
This sharing of program state interweaves the code interpreter and the language model simulator in a manner applicable to arbitrary interweaving, even control flow like \texttt{for}-loops and \texttt{if}-statements.
This continues until the entire code is run, and the answer is retrieved as the value of the variable named \texttt{answer}, or in case of irrecoverable errors, with the language model outputting \texttt{A: answer}.

To illustrate with a brief example, the code {\footnotesize\colorbox{highlight}{\texttt{answer = 0;}}} {\footnotesize\colorbox{highlight}{\texttt{answer += is\_sarcastic(`you don't say');}}} {\footnotesize\colorbox{highlight}{\texttt{answer += 1;}}} would be executed as follows:
(1) Python would execute the first line {\footnotesize \colorbox{execute-code-red}{\texttt{{answer = 0;}}}} and update the program state to {\footnotesize {\color{state-green}\texttt{\{answer = 0\}}}}, 
(2) Python would attempt to execute the second line and fail, and thus the LMulator would simulate the code   {\footnotesize \colorbox{execute-lm-purple}{\texttt{{answer += is\_sarcastic(`you don't say');}}}} by generating the program state {\footnotesize {\color{state-green}\texttt{\{answer = 1\}}}}, which would be updated in the program,
(3) Python would execute the last line {\footnotesize \colorbox{execute-code-red}{\texttt{{answer += 1;}}}} and update the program state to {\footnotesize {\color{state-green}\texttt{\{answer = 2\}}}},
(4) the answer would be retrieved as {\footnotesize \texttt{2}}.

\subsection{\algfullname Abilities}
\label{sec:abilities}

\algfullname has several attractive properties:
\begin{enumerate}
    \setlength\itemsep{-1mm}
    \item It enables code use in entirely new regimes, by combining the advantages of code with the powerful semantic and commonsense knowledge of language models, which can easily express rules that are challenging to express in code (e.g., which foods are fruits?). Such an ability may have benefits beyond reasoning problems and its flexibility enables executing expressive language, such as pseudocode.
    \item It leverages the ability of language models to code, a particular strength of recent language models due to the high quality data available.
    \item It inherits many of the benefits of reasoning code, both the formal yet expressive structure of code (e.g., Turing completeness) and powerful computational tools available to code (whether simply multiplying two numbers, calculating $\sqrt[5]{12121}$, or simulating physics).
    \item It inherits many of the benefits of techniques that reason via intermediate steps, such as Chain of Thought. These techniques enable the language model to use more computation when necessary to solve a problem as well as provide more interpretability.
\end{enumerate}
Empirically, we observe in Section \ref{sec:exp_lang} that these benefits results in significant improvements in reasoning performance over a variety of challenging tasks.

\section{Experimental Evaluation}
\label{sec:exp_lang}

We select challenging problems requiring varied types of reasoning, whether arithmetic, commonsense, or symbolic reasoning tasks, to answer the following questions:
\begin{enumerate}
    \setlength\itemsep{-1mm}
    \item How well does \algname perform across a variety of tasks?\label{q:overall}
    \item Which types of problems does \algname perform best?\label{q:problems}
    \item How does each aspect of \algname affect performance?\label{q:ablations}
    \item How does \algname scale with model size?\label{q:scaling}
    \item How does \algname perform as a general-purpose reasoner, with prompt examples from different problems rather than the same problem (which we term cross-task prompting)?\label{q:cross}
    \item How can \algname be used with instruction tuned chat models?\label{q:it}
    \item How robust \algname is against prompt variation?\label{q:robustness}
    \item Can \algname be applied beyond language reasoning tasks?\label{q:robotics}
\end{enumerate}

We first discuss the approaches, ablations, and baselines considered in Section~\ref{sec:exp_baselines}, then the tasks considered in Section~\ref{sec:exp_tasks}, and finally the results in Section~\ref{sec:exp_results}. 

\subsection{Baselines and Ablations}
\label{sec:exp_baselines}

We consider our main method to be \textbf{\algname (Interweave)}, also referred to as \textbf{\algname (Ours)}, though we also propose two variants with simpler implementation and modestly lower performance: \textbf{\algname (try Python except LM)} and \textbf{\algname (try Python except LM state)}. These two variants attempt to run the entire generated code with Python (rather than line by line) and if it fails, simulate the code execution with the LMulator, outputting a final answer or an intermediate state trace, respectively.
We also perform the following ablations, some of which are comparable to previous work as noted.
In \textbf{\algname (Python)} Python is used to run the entire generated code and if the code is not executable, it is marked as failure -- this can be thought of as a comparison to Program of Thoughts~\citep{chen2022program} or Program-aided language models~\citep{gao2023pal}.
We note that in many cases this baseline is particularly challenged, as writing executable code for some of the reasoning problems becomes nearly impossible (e.g., writing code to judge if a phrase is sarcastic), but one may focus on the results for Algorithmic only tasks for a more fair comparison.
In \textbf{\algname (LM)} the code is interpreted by an LMulator outputting the final answer, and in \textbf{\algname (LM state)} the code is interpreted by an LMulator outputting a state trace of intermediate steps -- this can be thought of as ScratchPad prompting for reasoning~\citep{nye2021show}. 
Note, the last two ablations do not leverage the Python interpreter.

We also compare against the following baselines.
In \textbf{Direct} question answering the LM simply responds to the question with a final answer.
In Chain of Thought prompting \textbf{(CoT)} the LM uses intermediate steps to solve the task; we use CoT as our standard prompt technique for the field of substep prompting~\citep{kojima2022large,zhou2022least} as prompts are readily available.


\subsection{Tasks}
\label{sec:exp_tasks}

We consider a subset of challenging tasks from BIG-Bench~\citep{srivastava2022beyond} called BIG-Bench Hard (BBH)~\citep{suzgun2022challenging} to ensure we are solving the most challenging tasks.
These tasks were specifically selected for their difficulty for language models and the datasets provides human-rater baselines and a set of Chain of Thought prompts.
The 23 tasks require semantic reasoning (e.g., ``Movie Recommendation''), numerical reasoning (e.g., ``Multi-Step Arithmetic''), and a combination of both (e.g., ``Object Counting'').
As such they enable us to study the efficacy of \algname across varied problems, not just those that coding is a natural fit for. Several prompts are shown in Figure~\ref{fig:qual_results}.
We also show results for the grade-school math (GSM8K) benchmark~\citep{cobbe2021training} in Section~\ref{sec:app_gsm}, although we find that these problems are primarily solved algorithmically alone through code.

These tasks are evaluated with \textbf{few-shot prompting}, whereby three examples from the same problem family are provided as context. 
We also introduce a new evaluation setting, \textbf{cross-task prompting}, whereby three examples of \textit{different} problems are provided as context.
As such, the language model has in-context examples of the \textit{format} of reasoning, but isn't provided explicit instructions on \textit{how} to reason.
We see this as an indicative signal for a general-purpose reasoner, which in many real-world applications (e.g., chatbots) would be asked to reason across a wide variety of tasks.

The models used herein include the OpenAI family of models: \texttt{text-ada-001}, \texttt{text-baggage-001}, \texttt{text-curie-001}, and \texttt{text-davinci-003} (in plots we denote these as a-1, b-1, c-1, and d-3). 
We also consider PaLM-2's code finetuned variant~\cite{chowdhery2022palm,anil2023palm}.
For instruction tuned models, we compare to recent variants of GPT (\texttt{gpt-3.5-turbo} and \texttt{gpt-4}) with the chat completion mode run in October 2023 and January 2024.
The results below are using the \texttt{text-davinci-003} model unless otherwise stated.

\subsection{Results}
\label{sec:exp_results}

\textbf{Question \ref{q:overall}: Overall Performance.}
The overall performance of \algname is shown in Figure~\ref{fig:overall} and Table~\ref{table:overall} (with full results in Table~\ref{table:all}).
We see that \algname outperforms other approaches, both in the number of tasks it exceeds the human baseline and in the overall amount that it exceeds the baseline. 
Indeed, CoC’s 84\% is SoTA to the best of our knowledge~\cite{gemini2023gemini}. In fact, when combined with \texttt{gpt-4}, CoC achieves 91\% (see Table~\ref{table:it_completion}).
In several tasks \algname vastly outperforms the human baseline and other methods, achieving nearly 100\% -- generally for these tasks the result is complicated in language but trivial in code (e.g., a task from multi-step arithmetic \texttt{Q:} $((-3 + 5 \times 8 \times -4) - (9 - 8 \times -7)) =$).
We also observe that CoT outperforms the human baseline on a number of tasks, while the Direct answer fares poorly.

\begin{table*}[htb]
  \centering
  \footnotesize
  \caption{
    Overall performance (\%) on BIG-Bench Hard with both few-shot prompting with a single task and cross-task. The delta compared to direct prompting is shown in parenthesis. 
  }
  \vspace{2mm}

  \begin{tabular}{@{}lccccccccccc@{}}
  \toprule
  & & \multicolumn{3}{c}{text-davinci-003} & \multicolumn{3}{c}{PaLM 2-S* (code variant~\cite{anil2023palm})} \\
  \cmidrule(lr){3-5} \cmidrule(lr){6-8}
 Prompt & Human & Direct & CoT & \algname (Ours) & Direct & CoT & \algname (Ours) \\  \midrule
Single task & 68 & 55 & 72 (+17) & 84 (+29) & 49 & 61 (+12) & 78 (+29) \\
Cross task & - & 50 & 55 (+5) & 61 (+11) & 45 & 47 (+2) & 47 (+2) \\  \bottomrule
  \end{tabular}
  \label{table:overall}
\end{table*}

\begin{figure*}[htb]
     \centering
         \centering
         \includegraphics[width=\textwidth]{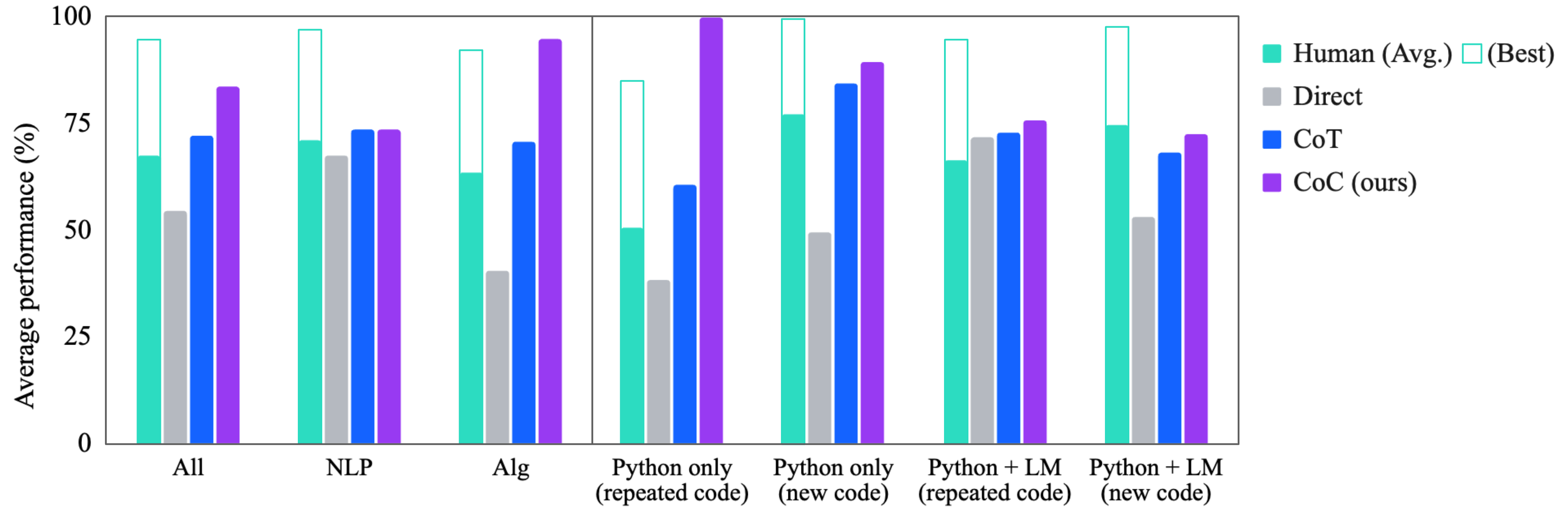}
        \vspace{-7mm}
        \caption{Average performance across different baselines grouped by task type, indicating the problem type and how \algname is generated \& executed.}
        \vspace{-3mm}
        \label{fig:by_task_type}
\end{figure*}

\textbf{Question \ref{q:problems}: Problem Type.}
Figure~\ref{fig:by_task_type} breaks the results down by problem type; the task labels are shown in Table~\ref{table:all}.
First, we isolate problems that are primarily algorithmic or primarily natural language (these categories were identified in \citep{suzgun2022challenging}).
We see that on algorithmic tasks, \algname performs particularly well, while on natural language tasks \algname performs on par with CoT. 
This is particularly encouraging, because one may expect these language oriented tasks to be a worse fit for code. The key is that our method offers the flexibility of using a LMulator to simulate the output of code execution, retaining the semantic reasoning capabilities of LMs for natural language problems. 

Figure~\ref{fig:by_task_type} additionally breaks the tasks down into categories that capture how different each question's response is and whether the code can be fully executed by Python (denoted Python only vs. Python + LM).
For some tasks within the benchmark, each question has the same code or Chain of Thought, with the only variation being the inputs -- in this case we say the code is (repeated code), and if not then it is denoted (new code).
As expected, we see that when the code is repeated and run by Python, \algname gets nearly 100\%, though these tasks (e.g., multi-step arithmetic) seem to be among the most challenging for the other baselines, including human raters.
The other categories are more challenging for \algname; however in each, we still see a benefit over baselines.

\begin{figure*}[htb]
     \centering
         \centering
         \includegraphics[width=\textwidth]{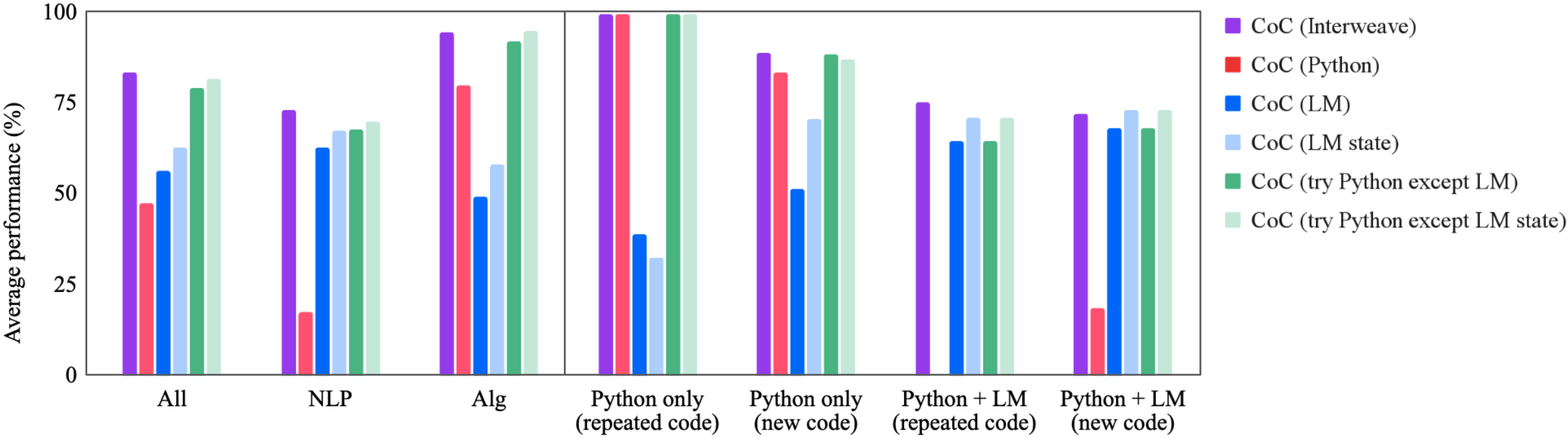}
         \vspace{-7mm}

        \caption{\algfullname ablations on average performance grouped by task type.}
        \label{fig:by_task_type_ablation}
\end{figure*}

\begin{figure*}[htb]
     \centering
     \begin{subfigure}[b]{0.3\textwidth}
         \centering
         \includegraphics[width=\textwidth]{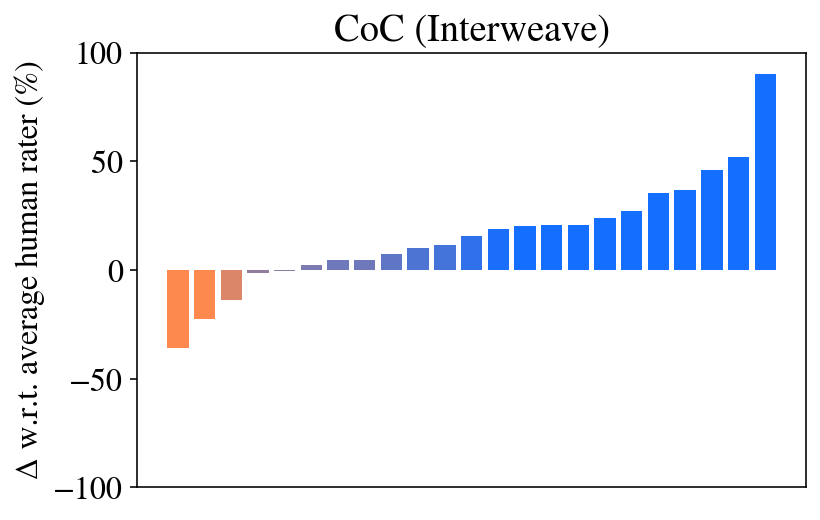}
         \label{fig:all_tasks_coc_interweave}
     \end{subfigure}
     \hfill
     \begin{subfigure}[b]{0.3\textwidth}
         \centering
         \includegraphics[width=\textwidth]{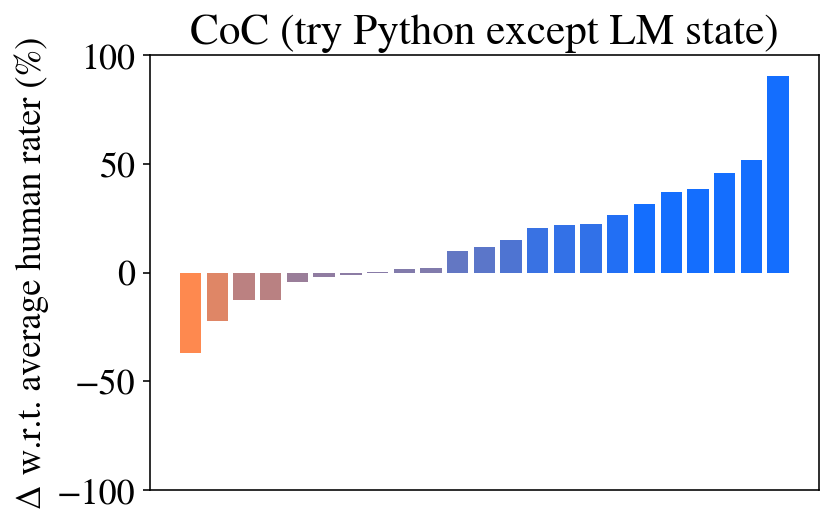}
         \label{fig:all_tasks_coc_try_except_llm_state}
     \end{subfigure}
     \hfill
     \begin{subfigure}[b]{0.3\textwidth}
         \centering
         \includegraphics[width=\textwidth]{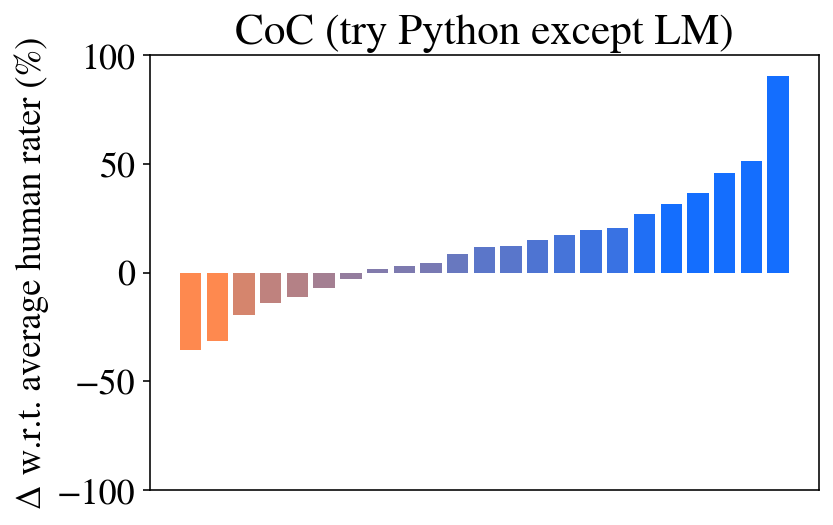}
         \label{fig:all_tasks_coc_try_except_llm}
     \end{subfigure}
     \hfill
     \begin{subfigure}[b]{0.3\textwidth}
         \centering
         \includegraphics[width=\textwidth]{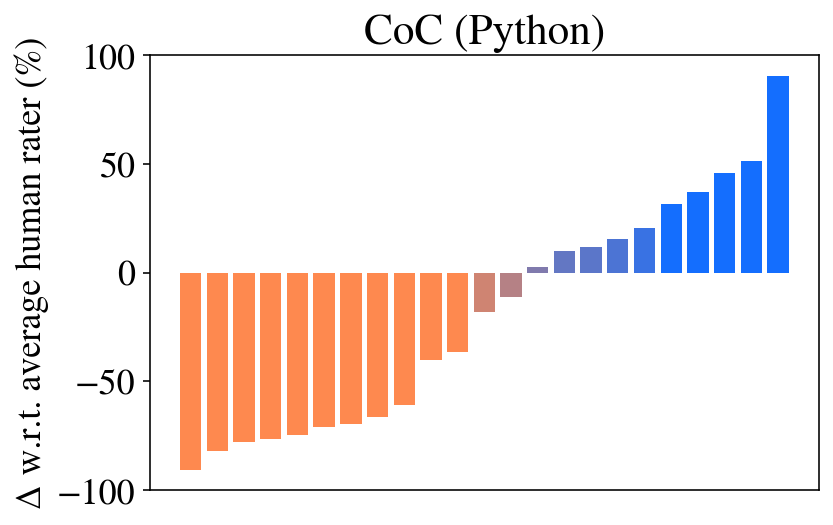}
         \label{fig:all_tasks_coc_python}
     \end{subfigure}
     \hfill
     \begin{subfigure}[b]{0.3\textwidth}
         \centering
         \includegraphics[width=\textwidth]{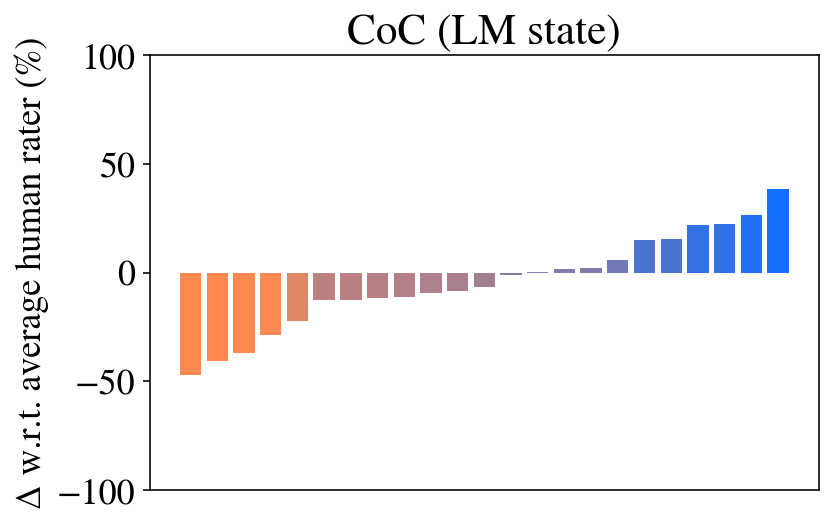}
         \label{fig:all_tasks_coc_llm_state}
     \end{subfigure}
     \hfill
     \begin{subfigure}[b]{0.3\textwidth}
         \centering
         \includegraphics[width=\textwidth]{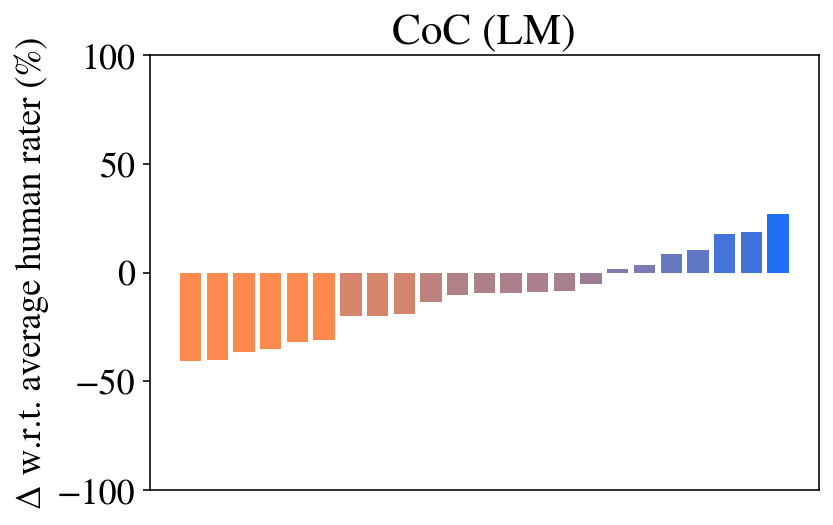}
         \label{fig:all_tasks_coc_llm}
     \end{subfigure}
    \vspace{-7mm}

    \caption{Results across all BIG-Bench Hard tasks compared to human baseline~\citep{srivastava2022beyond}. The tasks (x-axis) in each plot are sorted individually by performance. See Table~\ref{table:all} and Figure~\ref{fig:by_task_type_ablation} for a breakdown by task type.}
    \label{fig:all_tasks_ablation}
    \vspace{-3mm}

\end{figure*}

\textbf{Question \ref{q:ablations}: Ablations.}
Figures~\ref{fig:by_task_type_ablation} and~\ref{fig:all_tasks_ablation}, and Table~\ref{table:overall_ablations} show the ablations performed to motivate each aspect of \algfullname prompting.
As one may expect, the approaches that execute Python (\algname (Interweave, Python, try Python except LM, try Python except LM state)) achieve 100\% performance on several tasks -- if the code is correct, then the model will be correct every time.
However, the approach that relies on Python alone (\algname (Python)) performs poorly when applied to non-algorithmic tasks, failing almost all.
The \algname (Python) ablation is similar to recent works \citep{gao2023pal,chen2022program}, which show that if applied to numerical problems then code reasoning performs well.
\algname without the Python interpreter (\algname (LM, LM state)) too fares poorly, though we see that the step-by-step approach proposed in ScratchPad prompting~\citep{nye2021show} improves in each task.

We also show that ablations \algname (try Python except LM, try Python except LM state), in which \algname first tries to run the entire code with Python and if it fails simulates the code with an LM, perform quite well.
Again we see that maintaining a program state provides an improvement in performance.
With only minor degradations in performance observed, they are reasonable alternatives to the fully interweaved \algname for their simplicity.
Though we note, these ablations' performance would be much worse in cases where interweaving code and semantics is truly necessary -- for example, if we imagine a case where code is necessary to parse image inputs or to access an external database, but language is necessary to parse the results (see the robotics applications in Section~\ref{sec:exp_robot}).

\begin{table*}[htb]
  \centering
  \footnotesize
  \caption{
    Ablation overall performance (\%) with both few-shot prompting with a single task and cross-task. The delta compared to the full model (Interweave) is shown in parenthesis. 
  }
  \vspace{2mm}
  \begin{tabular}{@{}lcccccc@{}}
  \toprule
   & \multicolumn{5}{c}{\algfullname} \\
  \cmidrule(lr){2-7}
   & Interweave & try Python & try Python & Python & LM state & LM \\
   Prompt & & except LM state & except LM\\
  \midrule
Single task & 84 & 82 (-2) & 80 (-4) & 48 (-36) & 63 (-21) & 57 (-27) \\
Cross task & 61 & 57 (-4) & 60 (-1) & 35 (-26) & 49 (-12) & 50 (-11) \\
  \bottomrule
  \end{tabular}
  \label{table:overall_ablations}
  \vspace{-3mm}
\end{table*}

\begin{figure*}[htb]
     \centering
         \centering
         \includegraphics[width=\textwidth]{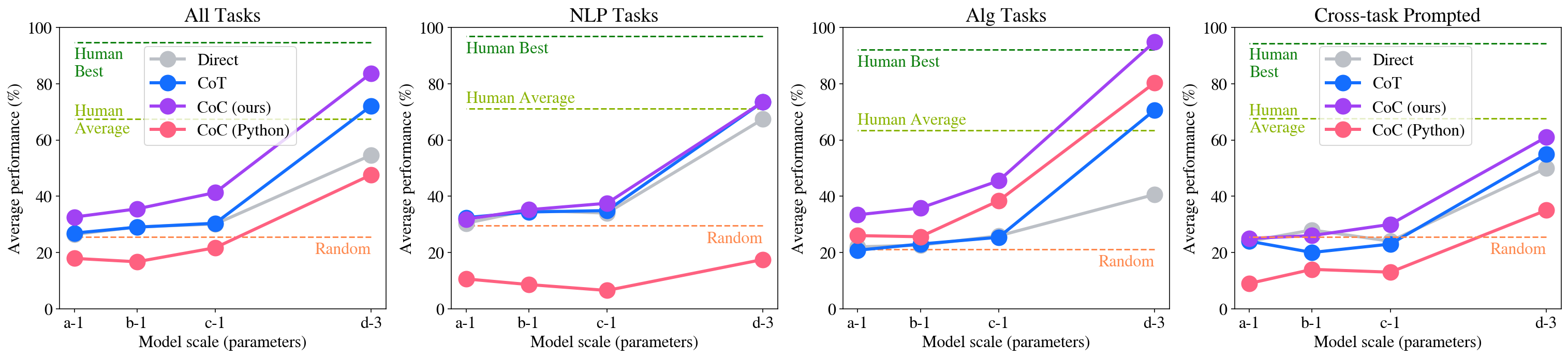}
\vspace{-7mm}
        \caption{Average performance with model scaling, from \texttt{text-ada-001} (smallest) to \texttt{text-davinci-003} (largest).}
        \label{fig:by_size}
\vspace{-2mm}
\end{figure*}

\textbf{Question \ref{q:scaling}: Scaling.}
Figure~\ref{fig:by_size} shows the performance of \algname across various model sizes. 
We observe that, similar to Chain of Thought prompting, the improvements of \algname increases as model size increases. In fact, for some of the algorithmic tasks, \algfullname even outperforms the best human raters (whom admittedly did not have access to code). Unlike Chain of Thought prompting, however, which only brings performance benefits for the largest model (d-3), \algname outperforms the direct question answering baseline also for smaller models (a-1, b-1, c-1), suggesting that it's easier for smaller models to output structured code as intermediate steps rather than natural languages.

\textbf{Question \ref{q:cross}: Cross-task Prompting.}
For cross-task prompting, we prompt the language models with a few examples from different problems. We see the performance drops for all methods in Figure~\ref{fig:by_size} and Table~\ref{table:overall}.
Despite this drop, \algname outperforms CoT and direct prompting at scale, nearly achieving human average performance.
This is a promising indication towards general purpose reasoning, in which a model does not expect to receive examples of similar problems in its prompt. 

\textbf{Question \ref{q:it}: Instruction Tuned Models.}
The reason why we chose \texttt{text-davinci-003}, a completion model, as our primary evaluation model, over more advanced instruction tuned models (\texttt{gpt-3.5-turbo} and \texttt{gpt-4}) is that the former is more amenable to few-shot prompting with examples, which is the main evaluation paradigm for BIG-Bench Hard. However, we still made our best attempt to evaluate our method with the  instruction tuned models using two different setups. The first is zero-shot prompting, where we directly prompt the models via the system message to output direct answers, chain of thoughts, or pseudocode/code (which we optionally execute with the python interpreter and feed back the results). The second is few-shot prompting, where we coerce the models to behave like completion models via the system message, and feed the few-shot examples as usual. In both cases, we demonstrated that \algname brings noticeable benefits with little modification needed. See Sec.~\ref{sec:it} for more details.

\textbf{Question \ref{q:robustness}: Robustness of \algfullname}
We showed that \algname is generally robust against prompt variation by evaluating with different prompts independently written by three annotators on the same set of problems. Specifically, we select four representative tasks from BIG-Bench Hard that require generation of new code (as opposed to repeated code). While the performance of individual tasks has some variance, the average performance across the four tasks only vary within a few percentage points. See Sec.~\ref{sec:robustness} for more details.

\textbf{Question \ref{q:robotics}: Beyond Language Reasoning}
We showed that \algname is well-suited for tasks that require both semantic and algorithmic reasoning beyond language reasoning, such as robotics. The unique advantage of \algname in robotics is that it interact seamlessly with the robot perception and control APIs via python code such as running object detectors or invoking parameterized robot skills, while performing semantic subtasks in an ``inline'' fashion (e.g. classifying what trash is compostable before picking them). When equipped with the necessary robot APIs, and a single example in the prompt to teach LMs the format, \algname can solve seven different robot manipulation tasks in the real world, showcasing generalization to new objects, languages and task domains. See Sec.~\ref{sec:exp_robot} for more details.

\section{Related Work}
\label{sec:rw}

\textbf{Language Model Reasoning}
The abilities and applications of language models have seen significant progress, 
due to their overall performance~\citep{chowdhery2022palm,touvron2023llama,radford2019language,gemini2023gemini} and emergent capabilities~\citep{wei2022emergent}, such as few-shot prompting~\citep{brown2020language} and abstract reasoning~\citep{wei2022chain}.
Perhaps most related to this work, a number of works have leveraged prompting to improve reasoning~\citep{dohan2022language}:
Chain of Thought~\citep{wei2022chain} proposes to break a task down into intermediate reasoning steps,
least-to-most~\citep{zhou2022least} proposes a series of increasingly simpler problems, and ScratchPad~\citep{nye2021show} proposes to maintain a trace of intermediate results for interpreting code (this first demonstrated the code simulation ability of LMs required for our LMulator).
Along these lines ``let's think step by step'' \citep{kojima2022large} uses a few key words to elicit such break downs (words that were later refined to ``Take a deep breath and work on this problem step-by-step'' in \cite{yang2023large}).
Beyond these, other approaches structure such step-by-step solutions into graphical structures~\citep{yao2023tree,besta2023graph},  plans~\citep{wang2023plan,ning2023skeleton}, or mixture of expert-based sampling~\citep{wang2022self,zhou2022mixture}.
\algname builds upon the intuition of these works, with the observation that \textit{code} is a formal, structured approach to breaking a problem down into sub-steps with many advantages beyond natural language alone.

\textbf{Language Model Tool Use}
Many recent works have proposed techniques for language models to use tools to respond to queries~\cite{mialon2023augmented}.
These tools have often been provided to the language model through prompting~\citep{cobbe2021training,khot2022decomposed,chowdhery2022palm,drori2022neural,yao2022react}, enabling tools like calculators for math problems, code interpreters, databases, or more.
These tools too can provide feedback on novel modalities~\citep{suris2023vipergpt,zeng2022socratic}. To expand the range of tools available, others have used external tool databases or finetuned language models~\citep{schick2023toolformer,qin2023toolllm,parisi2022talm,paranjape2023art}.
As tool interfaces vary, feedback from the tool too can improve performance~\cite{gou2023critic,zhou2023solving}.
In this work we leverage the expressibility and generality of full code as well as its structure, by treating it both as a tool and as a framework.

\textbf{Language Model Program Synthesis}
The ability of language models to code is well known and they have been applied as programming assistants~\citep{chen2021evaluating} and shown to be capable programmers on their own~\citep{austin2021program,li2022competition,nijkamp2022codegen}.
This ability has been applied to a variety of tasks outside of language alone, leveraging their ability to reason through code in new settings, such as robotics~\citep{liang2023code,singh2023progprompt}, embodied agents~\citep{wang2023voyager}, or vision~\citep{suris2023vipergpt}.
Others have specifically done so for reasoning, such as Program of Thoughts~\citep{chen2022program} and Program-aided Language Models~\citep{gao2023pal}, which generate code to solve numerical reasoning problems.
Herein, we focus on the interplay between writing code, running code, and language models simulating code,
thus enabling new regimes of language model code applications, such as semantic reasoning.

\section{Conclusions, Limitations, and Future Work}

We have proposed \algfullname, an approach towards reasoning with language models through writing code, and executing code either with an interpreter or with a language model that simulates the execution (termed herein an LMulator) if the code is not executable.
As such, \algname can leverage both the expressive structure of code and the powerful tools available to it.
Beyond this, by simulating the execution of non-executable code, \algname can apply to problems nominally outside the scope of code (e.g., semantic reasoning problems).
We have demonstrated that this approach outperforms baselines, and for some tasks even the best human raters, in a range of challenging language and numeric reasoning problems.

This work is not without its limitations.
First, generating and executing in two steps as well as interweaving code and language execution requires additional context length and computation time.
Second, though we have not seen any loss of performance for semantic tasks in aggregate, there are few tasks in which code doesn't help, e.g., the task Ruin Names, which asks whether an edit for a name is humorous.
Finally, our implementation to interweave LM and code is quite simple, tracking the program state in strings and parsing the strings into Python's built-in data types (e.g., dict, tuple). As our method stands now, the LM cannot modify custom Python objects while simulating code execution. In theory, however, it is doable as long as each of these Python objects have a serialization and deserialization method, e.g., using techniques like Protocol Buffers.  

There are many avenues for future work with \algname.
First, we believe that a unified code and language interpreter well combines the commonsense of language models with the analytical abilities, structure, and interpretability of code.
Such a technology can thus enable applications of code and code-like reasoning to novel problem regimes, beyond simple reasoning.
Second, we are interested in investigating the degree to which finetuning a language model to be an LMulator can benefit semantic code reasoning.
Third, we see evidence that reasoning through many pathways yields improvements, which is a promising step forward.
Finally, we believe this integration with code enables access to external modalities, such as vision or databases, and represents a interesting path for new applications (e.g., robotics, augmented reality).

\newpage

\section*{Impact Statement}
This paper presents work whose goal is to advance the field of Machine Learning. There are many potential societal consequences of our work, most of which are related to the usage of large language models (LLMs). One aspect of \algfullname that warrants further discussion is that \algname executes the output of LLMs using the Python interpreter as if they are always benign code. If deployed in the wild, however, \algfullname will need to install additional safeguards against potentially harmful code from LLMs that might be maliciously prompted, before running the code.  



\bibliography{example_paper}
\bibliographystyle{icml2024}

\newpage
\appendix
\onecolumn
\setcounter{table}{0}
\renewcommand{\thetable}{A\arabic{table}}
\setcounter{figure}{0}
\renewcommand{\thefigure}{A\arabic{figure}}

\section{Appendix}

\subsection{Quantitative results on language reasoning tasks}
\label{sec:app_quant_results}
Table~\ref{table:all} shows the full per-task results across ablations on BIG-Bench Hard (BBH) tasks, as well as broken down by task type and execution type. 

\begin{table}[htb]
  \centering
    \begin{adjustwidth}{.2in}{.2in}  
  \scriptsize
  \caption{
    Full results across ablations on BIG-Bench Hard (BBH) tasks. 
  }
  \vspace{2mm}
  \begin{tabular}{@{}b{3.8cm}B{0.7cm}B{0.7cm}B{0.7cm}B{0.7cm}B{0.7cm}B{0.7cm}B{0.7cm}B{0.7cm}B{0.7cm}B{0.7cm}B{0.7cm}@{}}
  \toprule
& \multicolumn{3}{c}{\tiny\citet{srivastava2022beyond}} \\
& \multicolumn{3}{c}{\tiny\citet{suzgun2022challenging}} & & &  \multicolumn{6}{c}{\algfullname} \\
  \cmidrule(lr){2-4} \cmidrule(lr){7-12} 
BIG-Bench Hard Task  & Rand. & Human (Avg.) & Human (Max) & Direct & CoT & Inter-weave & try Python except LM state & try Python except LM & Python & LM state & LM\\
\midrule
Boolean Expressions$^{\lambda+}$ & 50 & 79 & 100 & 88 & 89 & 100 & 100 & 100 & 100 & 95 & 90\\
Causal Judgement$^{\kappa*}$ & 50 & 70 & 100 & 64 & 64 & 56 & 57 & 63 & 0 & 57 & 60\\
Date Understanding$^{\kappa-}$ & 17 & 77 & 100 & 61 & 84 & 75 & 72 & 74 & 59 & 66 & 57\\
Disambiguation QA$^{\kappa/}$ & 33 & 67 & 93 & 70 & 68 & 71 & 67 & 68 & 0 & 67 & 68\\
Dyck Languages$^{\lambda+}$ & 1 & 48 & 100 & 6 & 50 & 100 & 100 & 99 & 99 & 1 & 7\\
Formal Fallacies$^{\kappa*}$ & 25 & 91 & 100 & 56 & 56 & 55 & 54 & 55 & 0 & 54 & 56\\
Geometric Shapes$^{\lambda+}$ & 12 & 54 & 100 & 48 & 66 & 100 & 100 & 100 & 100 & 13 & 44\\
Hyperbaton$^{\kappa/}$ & 50 & 75 & 100 & 63 & 64 & 98 & 62 & 55 & 0 & 62 & 55\\
Logical Deduction$^{\lambda*}$ & 23 & 40 & 89 & 49 & 66 & 68 & 79 & 57 & 0 & 79 & 58\\
Movie Recommendation$^{\kappa/}$ & 25 & 61 & 90 & 85 & 81 & 80 & 83 & 80 & 0 & 83 & 79\\
Multi-Step Arithmetic$^{\lambda+}$ & 0 & 10 & 25 & 0 & 48 & 100 & 100 & 100 & 100 & 0 & 1\\
Navigate$^{\lambda*}$ & 50 & 82 & 100 & 58 & 94 & 86 & 84 & 68 & 0 & 84 & 68\\
Object Counting$^{\lambda-}$ & 0 & 86 & 100 & 30 & 82 & 96 & 98 & 98 & 98 & 57 & 50\\
Penguins in a Table$^{\kappa-}$ & 0 & 78 & 100 & 62 & 82 & 90 & 88 & 90 & 88 & 71 & 59\\
Reasoning about Colored Objects$^{\kappa-}$ & 12 & 75 & 100 & 64 & 87 & 78 & 74 & 78 & 64 & 64 & 70\\
Ruin Names$^{\kappa/}$ & 25 & 78 & 100 & 76 & 70 & 55 & 56 & 46 & 0 & 56 & 47\\
Salient Translation Error Detection$^{\kappa/}$ & 17 & 37 & 80 & 66 & 61 & 58 & 63 & 64 & 0 & 63 & 64\\
Snarks$^{\kappa/}$ & 50 & 77 & 100 & 70 & 71 & 76 & 76 & 66 & 0 & 76 & 66\\
Sports Understanding$^{\kappa/}$ & 50 & 71 & 100 & 72 & 96 & 91 & 93 & 75 & 0 & 93 & 74\\
Temporal Sequences$^{\lambda*}$ & 25 & 91 & 100 & 38 & 60 & 98 & 93 & 99 & 93 & 93 & 99\\
Tracking Shuffled Objects$^{\lambda-}$ & 23 & 65 & 100 & 25 & 72 & 100 & 96 & 96 & 96 & 71 & 24\\
Web of Lies$^{\lambda-}$ & 50 & 81 & 100 & 54 & 100 & 97 & 96 & 96 & 97 & 96 & 50\\
Word Sorting$^{\lambda+}$ & 0 & 63 & 100 & 51 & 50 & 99 & 100 & 99 & 100 & 54 & 54\\
\midrule
Task Averages \\
NLP Task (avg)$^{\kappa}$ & 30 & 71 & 97 & 67 & 74 & 74 & 70 & 68 & 18 & 68 & 63\\
Algorithmic Task (avg)$^{\lambda}$ & 21 & 64 & 92 & 41 & 71 & 95 & 95 & 92 & 80 & 58 & 50\\
All Tasks (avg)$^{}$ & 26 & 68 & 95 & 55 & 72 & 84 & 82 & 80 & 48 & 63 & 57\\
\midrule
Execution Type \\
Python exec (same program)$^{+}$ & 13 & 51 & 85 & 38 & 61 & 100 & 100 & 100 & 100 & 33 & 39\\
Python exec (different program)$^{-}$ & 17 & 77 & 100 & 49 & 84 & 89 & 87 & 89 & 84 & 71 & 52\\
LM exec (same program)$^{/}$ & 36 & 66 & 95 & 72 & 73 & 76 & 71 & 65 & 0 & 71 & 65\\
LM exec (different program)$^{*}$ & 35 & 75 & 98 & 53 & 68 & 72 & 73 & 68 & 19 & 73 & 68\\
  \bottomrule
  \end{tabular}
    \scriptsize
    $\lambda$ denotes an algorithmic task and $\kappa$ denotes an NLP task (with categories outlined in \citet{suzgun2022challenging}). 
    $+$ denotes a task where the code between prompts is repeated and can be executed by Python, 
    $-$ denotes a task where the code between prompts must change and can be executed by Python, 
    $/$ denotes a task where the code between prompts is repeated and must be executed by the LM, and 
    $*$ denotes a task where the code between prompts must change and must be executed by the LM.
  \label{table:all}
      \end{adjustwidth}
\end{table}

\subsection{Quantitative results on the GSM8K Benchmark}
\label{sec:app_gsm}
Table~\ref{table:gsm} shows results on the the grade-school math benchmark (GSM8K)~\cite{cobbe2021training} with direct prompting, Chain of Thought, and \algfullname.
We find that \algname generally outperforms CoT and Direct prompting. Since these tasks are primarily algorithmic and are solved by Python alone, all \algfullname variants that use Python achieve the same performance  -- also the same performance shown in Program of Thoughts~\cite{chen2022program}.

\begin{table}[bth]
  \centering
  \footnotesize
  \caption{
    GSM8K~\cite{cobbe2021training} performance (\%) with both few-shot prompting with a single task and cross-task. The delta compared to direct prompting is shown in parenthesis. 
  }
  \vspace{2mm}
  \begin{tabular}{@{}lcccccccc@{}}
  \toprule
   & & & \multicolumn{5}{c}{\algfullname} \\
  \cmidrule(lr){4-9}
Prompt & Direct & CoT & Interweave & try Python & try Python & Python only & LM state & LM only \\
& & & & except LM state & except LM  \\
\midrule
Single task & 16 & 63 (47) & 71 (55) & 72 (56) & 71 (55) & 71 (55) & 45 (29) & 22 (6) \\
Cross task & 14 & 55 (41) & 60 (46) & 60 (46) & 60 (46) & 60 (46) & 41 (27) & 16 (2) \\
  \bottomrule
  \end{tabular}
  \label{table:gsm}
\end{table}

\subsection{Qualitative results on language reasoning tasks}
\label{sec:app_qual_results}
Figure~\ref{fig:qual_results} shows the model outputs for a few reasoning tasks from BIG-Bench Hard (BBH) and Figure~\ref{fig:intro_holiday} shows a demonstrative example of date reasoning. These examples are selected to highlight the interweaving execution of the Python interpreter and the LMulator.

\begin{figure}[h]
\centering
\begin{subfigure}{.5\linewidth}
\caption{Movie Recommendation\vspace{-5pt}}
  \setlength{\fboxsep}{2pt}
  \flushright
  \noindent\fbox{\parbox{0.93\linewidth}{\tiny{\texttt{{
  \\
  {Q: Find a movie similar to Batman, The Mask, The Fugitive, Pretty Woman:\\
  Options:\\
  (A) The Front Page\\
  (B) Maelstrom\\
  (C) The Lion King\\
  (D) Lamerica}\\ \\
\hlruncode{options = \{"The Front Page": "(A)", "Maelstrom": "(B)", "The Lion King": "(C)", "Lamerica": "(D)"\}\\
{\color{state-green}delta state: \{options = \{`The Front Page': ..., ...\}\}}}
\hlruncode{query\_movies = list(options.keys())\\
{\color{state-green}delta state: \{query\_movies = [`The Front Page', ...]\}}}
\hlruncode{ref\_movies = ["Batman", "The Mask", "The Fugitive", "Pretty Woman"]\\
{\color{state-green}delta state: \{ref\_movies = [`Batman', ..., `Pretty Woman']\}}}
\hlrunlm{ref\_movie\_infos = get\_movie\_genre\_and\_year(ref\_movies, ret\_type="list[tuple[str, int]]")\\
{\color{state-green}delta state: \{ref\_movie\_infos = [(`action', 1989), ...]\}}}
\hlrunlm{most\_similar\_movie = get\_most\_similar\_movie(query\_movies, ref\_movies, ref\_movie\_infos, ret\_type="str")\\
{\color{state-green}delta state: \{most\_similar\_movie = `The Lion King'\}}}
\hlruncode{answer = options[most\_similar\_movie] if most\_similar\_movie in options else None\\
{\color{state-green}delta state: \{answer = `(C)'\}}}
  }}}}}\\
\end{subfigure}
\hspace{-5pt}
\begin{subfigure}{.5\linewidth}
\caption{Hyperbaton\vspace{-5pt}}
  \setlength{\fboxsep}{2pt}
  \flushright
  \noindent\fbox{\parbox{0.93\linewidth}{\tiny{\texttt{{
  \\
  {Q: Which sentence has the correct adjective order:\\
  Options:\\
  (A) small grey square cloth Japanese shoe\\
  (B) small square grey Japanese cloth shoe"}\\ \\
\hlruncode{import numpy as np\\
{\color{state-green}delta state: \{\}}}
\hlruncode{options = \{"(A)": "small grey square cloth Japanese shoe", "(B)": "small square grey Japanese cloth shoe"\}\\
{\color{state-green}delta state: \{options = \{`(A)': ..., `(B)': ...\}\}}}
\hlruncode{priority = \{"opinion": 1, "size": 2, "age": 3, "shape": 4, "color": 5, "origin": 6, "material": 7, "purpose": 8\}\\
{\color{state-green}delta state: \{priority = \{`opinion': .., .., `purpose': ..\}\}}}
\hlruncode{valid\_types = list(priority.keys())\\
{\color{state-green}delta state: \{valid\_types = [`opinion', ..., `purpose']\}}}
\hlruncode{scores = []\\
{\color{state-green}delta state: \{scores = []\}}}
\hlruncode{for option, sentence in options.items():\\
{\color{state-green}delta state: \{option, sentence = `(A)', `small ... shoe'\} \\ \# updated for each loop}}
\hlruncode{\quad adjs = sentence.split(" ")[:-1]\\
{\color{state-green}delta state: \{adjs = [`small', `grey', `square', `cloth']\}}}
\hlrunlm{\quad order = [priority[get\_adjective\_type(adj, valid\_types, ret\_type=str)] for adj in adjs]\\
{\color{state-green}delta state: \{order = [2, 5, 4, 6]\}}}
\hlruncode{\quad scores.append([order[i+1] > order[i] for i in range(len(order) - 1)].count(True))\\
{\color{state-green}delta state: \{scores = [2]\}}}
\hlruncode{answer = list(options.keys())[np.argmax(scores)]\\
{\color{state-green}delta state: \{answer = `(B)'\}}}
  }}}}}
\end{subfigure}
\vspace{2mm}
\hrule
\vspace{2mm}

\begin{subfigure}{.5\linewidth}
\caption{Logical Deduction\vspace{-5pt}}
  \setlength{\fboxsep}{2pt}
  \flushright
  \noindent\fbox{\parbox{0.93\linewidth}{\tiny{\texttt{{
  \\
  {Q: The following paragraphs each describe a set of three objects arranged in a fixed order. The statements are logically consistent within each paragraph. On a shelf, there are three books: a green book, a red book, and a blue book. The red book is the rightmost. The blue book is to the right of the green book.\\
  Options:\\
  (A) The green book is the leftmost\\
  (B) The red book is the leftmost\\
  (C) The blue book is the leftmost}\\\\
\hlruncode{options = \{"green": "(A)", "red": "(B)", "blue": "(C)"\}\\
{\color{state-green}delta state: \{options = \{`green': ..., ..., `blue': ...\}\}}}
\hlruncode{order\_info = "left to right"\\
{\color{state-green}delta state: \{order\_info = `left to right'\}}}
\hlruncode{full\_order = [None, None, None]\\
{\color{state-green}delta state: \{full\_order = [None, None, None]\}}}
\hlruncode{partial\_order = []\\
{\color{state-green}delta state: \{partial\_order = []\}}}
\hlruncode{full\_order[-1] = "red"\\
{\color{state-green}delta state: \{full\_order = [None, None, `red']\}}}
\hlruncode{partial\_order.append(("green", "blue"))\\
{\color{state-green}delta state: \{partial\_order = [(`green', `blue')]\}}}
\hlrunlm{full\_order = generate\_full\_order(full\_order, partial\_order, ret\_type=list)\\
{\color{state-green}delta state: \{full\_order = [`green', `blue', `red']\}}}
\hlruncode{query = "leftmost"\\
{\color{state-green}delta state: \{query = `leftmost'\}}}
\hlrunlm{result = query\_result(order\_info, full\_order, query, ret\_type=str)\\
{\color{state-green}delta state: \{result = `green'\}}}
\hlruncode{answer = options[result] if result in options else None\\
{\color{state-green}delta state: \{answer = `(A)'\}}}
  }}}}}\\
\end{subfigure}
\hspace{-5pt}
\begin{subfigure}{.5\linewidth}
\caption{Disambiguation QA\vspace{-5pt}}
  \setlength{\fboxsep}{2pt}
  \flushright
  \noindent\fbox{\parbox{0.93\linewidth}{\tiny{\texttt{{
  \\
  {Q: In the following sentences, explain the antecedent of the pronoun (which thing the pronoun refers to), or state that it is ambiguous.\\
Sentence: The homeowner asked the inspector if the house they had purchased was structurally sound.\\
Options:\\
(A) The homeowner had purchased\\
(B) The inspector had purchased\\
(C) Ambiguous}\\ \\
\hlruncode{context = "The homeowner asked the inspector if the house they had purchased was structurally sound."\\
{\color{state-green}delta state: \{context = `The homeowner asked ... sound.'\}}}
\hlruncode{pronoun = "they"\\
{\color{state-green}delta state: \{pronoun = `they'\}}}
\hlruncode{a = "homeowner"\\
{\color{state-green}delta state: \{a = `homeowner'\}}}
\hlruncode{b = "inspector"\\
{\color{state-green}delta state: \{b = `inspector'\}}}
\hlruncode{version\_a = "The homeowner asked the inspector if the house the homeowner had purchased was structurally sound."\\
{\color{state-green}delta state: \{version\_a = `The homeowner asked ... sound.'\}}}
\hlruncode{version\_b = "The homeowner asked the inspector if the house the inspector had purchased was structurally sound."\\
{\color{state-green}delta state: \{version\_b = `The homeowner asked ... sound.'\}}}
\hlrunlm{valid\_a = can\_pronoun\_refer\_to\_noun(pronoun=pronoun, noun=a, full\_sentence=version\_a, ret\_type=bool)\\
{\color{state-green}delta state: \{valid\_a = True\}}}
\hlrunlm{valid\_b = can\_pronoun\_refer\_to\_noun(pronoun=pronoun, noun=b, full\_sentence=version\_b, ret\_type=bool)\\
{\color{state-green}delta state: \{valid\_b = False\}}}
\hlruncode{if valid\_a and not valid\_b:\\
{\color{state-green}delta state: \{\}}}
\hlruncode{\quad answer = "(A)"\\
{\color{state-green}delta state: \{answer = `(A)'\}}}
\hlruncode{elif valid\_b and not valid\_a:}
\hlruncode{\quad answer = "(B)"}
\hlruncode{else:}
\hlruncode{\quad answer = "(C)"}
  }}}}}
\end{subfigure}
\caption{Model outputs for a few reasoning tasks from BIG-Bench Hard (BBH). We observe that \algname can apply to a wide variety of complex reasoning tasks that involve both semantic and numeric reasoning. \colorbox{execute-code-red}{Red} highlight indicates LM generated code being executed by the Python interpreter, and \colorbox{execute-lm-purple}{purple} highlight indicates LM simulating the code execution.}
\label{fig:qual_results}
\end{figure}

\begin{figure}[h]
\centering
\begin{subfigure}{.45\linewidth}
\caption*{Direct answer only}
\vspace{-2mm}
  \setlength{\fboxsep}{3pt}
  \flushleft
  \noindent\fbox{\parbox{0.93\linewidth}{\scriptsize{\texttt{{
  \\
  {Q: What holiday is 314 days after Valentine's Day in 2024?}\\ 
  \hlgen{\textbf{A: Christmas (35\%, \redcross), New Year's (15\%, \redcross), Other (45\%, \redcross), Christmas Eve (0\%, \greentick)}}}\\
  }}}}\\
~
\caption*{Chain of Thought}
\vspace{-2mm}
  \setlength{\fboxsep}{3pt}
  \flushleft
  \noindent\fbox{\parbox{0.93\linewidth}{\scriptsize{\texttt{{
  \\
  {Q: What holiday is 314 days after Valentine's Day in 2024? Let's think step by step.}\\ 
  \hlgen{Step 1: Count the days in February after Valentine's Day: ... \\\textbf{A: Christmas Eve (35\%, \greentick), Christmas (5\%, \redcross), Nov X (20\%, \redcross), Jan X (20\%, \redcross), Other (20\%, \redcross)}}\\
  }}}}}\\
  \label{fig:intro_holiday_generate}
  \vspace{-20.5mm}
\end{subfigure}%
\begin{subfigure}{.53\linewidth}
\caption*{\algfullname}
\vspace{-2mm}
  \setlength{\fboxsep}{3pt}
  \flushright
  \noindent\fbox{\parbox{0.93\linewidth}{\scriptsize{\texttt{{
  \\
  {Q: What holiday is 314 days after Valentine's Day in 2024?}\\ \\
  \hlruncode{\linenumber{1}from datetime import date, timedelta \\{\color{state-green}delta\_state: \{\}}}\\
  \hlrunlm{\textcolor{gray}{2}\ \ \ \ day1 = get\_valentines\_day\_date(2024) \\{\color{state-green}delta\_state: \{day1 = date(year=2024, month=2, day=14)\}}}\\
  \hlruncode{\textcolor{gray}{3}\ \ \ \ day2 = day1 + timedelta(days=314) \\{\color{state-green}delta\_state: \{day2 = date(year=2024, month=12, day=24)\}}}\\
  \hlrunlm{\textcolor{gray}{4}\ \ \ \ answer = get\_holiday(day2) \\{\color{state-green}delta\_state: \{answer = `Christmas Eve'\}}}\\
  \\ \textbf{{A: Christmas Eve (100\%, \greentick)}}\\
  }}}}}\\
\end{subfigure}%
\caption{
     A demonstrative example of how \algfullname generates code and reasons through an LM-augmented code emulator. Lines evaluated with Python are in \colorbox{execute-code-red}{red} and with an LM are in \colorbox{execute-lm-purple}{purple}. The chain of thought and direct answers were evaluated with \texttt{gpt-4} in October 2023, and we note the current model (as of December 2023) writes code to solve this problem and gets the same solution as \algfullname.
     }
\vspace{-1.2em}
\label{fig:intro_holiday}
\end{figure}

\subsection{Instruction Tuned Models}
\label{sec:it}
Since most of the results presented in our main paper are using \texttt{text-davinci-003}, a completion model that is particularly amenable to few-shot prompting, one may wonder how \algname can be used with instruction tuned models, like \texttt{gpt-4}~\cite{openai2023gpt4}. Figuring out ways to elicit the desired behavior of \algname from these instruction tuned models (i.e. writing code/pseudocode to solve problems) is non-trivial. We conduct two additional experiments below as our best effort to shed some light on this subject.

\textbf{Zero-shot prompting.} We directly prompt the models with instructions to elicit the desired reasoning approaches. Note that we do not provide few-shot examples in the prompt (hence ``zero-shot'').
For the baselines, we ask the model to ``directly answer'' (Direct) or ``think step by step'' (CoT).
For \algname variants, we ask the model to ``write python code to help solve the problem, if it's helpful''.
If a program is written, we either run the code with a Python interpreter and then feed the result (or the error message if execution fails) back to the model to determine a final answer (\algname (Python)), or ask the model to simulate the output of code execution as a LMulator (\algname (LM)).
The \algname (Python) baseline can be thought of as a comparison to an LM with Python tool use.

Table~\ref{table:it} shows the performance of each.
With \texttt{gpt-3.5-turbo}, both CoT and \algname (Python) show benefits over direct prompting, although both are strongly outperformed by \algname (Interweave).
With \texttt{gpt-4}, despite the considerable model strength advantage over \texttt{text-davinci-003}, \algname (Interweave) still outperforms, though the gap is narrower.
Admittedly, \algname (Interweave) is prompted with three examples whereas the other two are not.


\vspace{-3mm}
\begin{table*}[htb]
  \centering
  \footnotesize
  \caption{
    Comparisons with instruction tuned models in the chat interface, with and without tool use, denoted as \algname (Python) and \algname (LLM) respectively. The delta compared to \algname with text-davinci-003 is shown in parenthesis. In this experiment, the prompts for \texttt{gpt-3.5-turbo} and \texttt{gpt-4} only contain a generic, shared system message and do not contain few-shot examples.
  }
  \vspace{2mm}
  \begin{tabular}{@{}ccccccccc@{}}
  \toprule
  \multicolumn{1}{c}{text-davinci-003} & \multicolumn{4}{c}{gpt-3.5-turbo} & \multicolumn{4}{c}{gpt-4} \\
  \cmidrule(lr){1-1} \cmidrule(lr){2-5} \cmidrule(lr){6-9}
\algname & Direct & CoT & \algname & \algname & Direct & CoT & \algname & \algname \\
(Interweave) &  &  & (Python) & (LM) & &  & (Python)  & (LM)\\
\midrule
84 & 51 (-33) & 56 (-28) & 56 (-28) & 45 (-39) & 70 (-14) & 78 (-6) & 82 (-2) & 75 (-9)\\
  \bottomrule
  \end{tabular}
  \label{table:it}
\end{table*}

\textbf{Few-shot prompting.} We attempt to coerce the instruction tuned models to behave like completion models by using the following system message: ``Pretend you are a completion model that is prompted with three examples. You should follow the pattern of these examples strictly. At the end of your reply, you should always output an answer''. In the user message, we include three examples from the same (single task) or different (cross task) task domains, as well as the query question, exactly the same as how we evaluated the completion models.

Table~\ref{table:it_completion} shows that \algname still brings a sizable performance gain over the Direct and CoT baselines. With \texttt{gpt-4}, the gap is again narrower mainly because the base model already performs quite well and leaves little room for improvement. This experiment suggests a viable way to combine the strength of \algname with that of more advanced instruction tuned models like \texttt{gpt-4}.  

\begin{table*}[htb]
  \centering
  \footnotesize
  \caption{
    Applying \algname to instruction tuned models in the chat interface, while coercing them to behave like completion models. The delta compared to direct prompting is shown in parenthesis. In this experiment, the prompts contains a generic, shared system message that asks LMs to behave like completion models, and also three examples from the same or different task domains at the beginning of the user message. as before.
  }
  \vspace{2mm}

  \begin{tabular}{@{}lccccccccccc@{}}
  \toprule
  & \multicolumn{3}{c}{gpt-3.5-turbo} & \multicolumn{3}{c}{gpt-4} \\
  \cmidrule(lr){2-4} \cmidrule(lr){5-7}
 Prompt & Direct & CoT & \algname & Direct & CoT & \algname \\  \midrule
Single task & 47 & 73 (+26) & 79 (+32) & 69 & 88 (+19) & 91 (+22) \\
Cross task & 47 & 60 (+13) & 61 (+14) & 67 & 81 (+14) & 84 (+17) \\  \bottomrule
  \end{tabular}
  \vspace{-2mm}

  \label{table:it_completion}
\end{table*}

\subsection{Robustness of \algfullname}
\label{sec:robustness}
Similar to Chain of Thought prompts, \algfullname prompts can also come with various forms: e.g. different ways of function decomposition, coding styles, variable names, reasoning pathways, and so on. In this section, we want to analyze whether \algname is robust against variation across prompts.

We invite three annotators that are familiar with Python to write \algname prompts for four representative tasks in BIG-Bench Hard. We select these four tasks because they all require generation of new code (as opposed to repeated code) during test time. As before, for single task evaluation, we prompt LMs with three examples from the same task domain, whereas for cross task evaluation, we prompt LMs with three examples from different task domains (one from each of the other three tasks).  

Results in Table~\ref{table:prompt_robustness} show that our method is robust against prompt variation and doesn't require extensive prompt engineering.

\begin{table*}[ht]
  \centering
  \footnotesize
  \caption{
    Performance variation across prompts written independently by different authors for four representative tasks in BIG-Bench Hard. Our results that \algname is generally robust against prompt variation, allowing for different coding styles and reasoning logic in the few-shot prompts.
  }
  \vspace{2mm}

  \begin{tabular}{@{}lcccccccccccc@{}}
  \toprule
Prompt & Prompt Annotator & Date Understanding & Logical Deduction & Object Counting & Penguins in a Table & Average \\  \midrule
Single task & A & 73 & 64 & 92 & 78 & 77 \\
& B & 68 & 54 & 95 & 88 & 76 \\ 
& C & 69 & 43 & 90 & 89 & 73 \\ \midrule
Cross task & A & 41 & 33 & 67 & 76 & 54 \\
& B & 48 & 29 & 78 & 88 & 61 \\ 
& C & 60 & 30 & 76 & 64 & 57 \\ 
\bottomrule
  \end{tabular}
  \label{table:prompt_robustness}
\end{table*}


\subsection{Robotics Applications}
\label{sec:exp_robot}
Downstream applications such as robotics are well fit for \algname as robotics tasks require semantic reasoning and algorithmic reasoning, as well as interfacing with other APIs through code (such as control or perception APIs~\cite{liang2023code}) and with users through natural language. 
For example, given a task like ``sort the fruits by size'', the robot must reason over which items are fruits, sort them by size, and then connect those decisions to actions executable on the robot.
\algname (Interweave) is able to solve these challenges with the Python interpreter and the LMulator at runtime, while allowing for more 
interpretability and fine-grained control of the robot policies.  

\textbf{Environment and Robot Setup.}
Our environment is a tabletop with small objects (containers, toys, etc) and a UR5 robot arm equipped with a vacuum gripper and a wrist-mounted RGB-D camera. 
For the purpose of our experiments, the available perception API is \texttt{detect\_objects()}, which returns a list of detected objects (probabilities, labels, bounding boxes and segmentation masks) from the wrist camera. 
This API is implemented with first querying GPT-4V~\citep{openai2023gpt4} for a list of objects, and then using Grounding-SAM~\citep{kirillov2023segany, liu2023grounding} to localize them. The available control API is \texttt{pick\_place(obj1, obj2)}, which is a scripted primitive skill that picks up \texttt{obj1} and places it on top of \texttt{obj2}. There is also a text-to-speech API \texttt{say(sentence)} that allows the robot to communicate with the user.

\textbf{Experimental Setup.}
We evaluate with a number of tabletop pick-and-place robotics tasks that involve semantic reasoning. 
For few-shot prompting, we include a single example: ``Serve a meal that follows the user's dietary restrictions'', so that the language model understands the expected structure as well as the available robot APIs. During test time, we query the model with each of the following instructions.
\begin{enumerate}
    \setlength\itemsep{-1mm}
    \item ``Pack a lunch box for someone who is on a vegan diet.''
    \item ``Assemble a sandwich for someone who is vegetarian.''
    \item ``Gather ingredients for a peanut butter sandwich in a plate.''
    \item ``Prepare 西红柿炒蛋\  in the pot.'' (interleaving English and Chinese on purpose)
    \item ``Place all paper-made objects in the grass-colored container.''
    \item ``Sort the objects on the table into the compost bin and the recycle bin.''
    \item ``My steak is too bland. Can you help?''
\end{enumerate}

\textbf{Results.}
With a single example in our prompt, we see that our model is able to generalize to new objects, languages, and task domains (see Figure~\ref{fig:robot_exp} and an example trajectory in Figure~\ref{fig:robot_traj}).
Note that for these robotics tasks, unlike the previous language reasoning tasks, our main method \algname (Interweave) is the only capable approach, as the code requires line-by-line interplay between the Python interpreter execution (robot APIs) and the LMulator (commonsense QA like \texttt{is\_compostable}).

Figure~\ref{fig:robot_exp} shows the one-shot prompt as well as the model outputs and how they are executed for a few test instructions.

\begin{figure}[h]
\centering
\begin{subfigure}{.5\linewidth}
\caption{Given Prompt\vspace{-5pt}}
  \setlength{\fboxsep}{2pt}
  \flushright
  \noindent\fbox{\parbox{0.93\linewidth}{\tiny{\texttt{{
  \\
  {Q: Serve a meal that follows the user's dietary restrictions.}\\ \\
\hlruncode{restrictions = input("Do you have any dietary restrictions?")}
\hlruncode{\_, \_, objs, \_ = robot.detect\_objects()}
\hlruncode{container\_obj = None}
\hlruncode{for obj in objs:}
\hlrunlm{\quad is\_match\_description = match\_description(obj, "food container", ret\_type="bool")}
\hlruncode{\quad if is\_match\_description:}
\hlruncode{\quad \quad container\_obj = obj}
\hlruncode{\quad \quad break}
\hlruncode{if container\_obj is None:}
\hlruncode{\quad robot.say("No food container found.")}
\hlruncode{else:}
\hlruncode{\quad for obj in objs:}
\hlruncode{\quad \quad if obj == container\_obj:}
\hlruncode{\quad \quad \quad continue}
\hlrunlm{\quad \quad is\_follow\_restrictions = follow\_restrictions(obj, restrictions, ret\_type="bool")}
\hlruncode{\quad \quad if is\_follow\_restrictions:}
\hlruncode{\quad \quad \quad robot.pick\_place(obj, container\_obj)}
  }}}}}
\end{subfigure}
\hspace{-5pt}
\begin{subfigure}{.5\linewidth}
\caption{Novel Objects\vspace{-5pt}}
  \setlength{\fboxsep}{2pt}
  \flushright
  \noindent\fbox{\parbox{0.93\linewidth}{\tiny{\texttt{{
  \\
  {Q: Sort the objects on the table into the compost bin and the recycle bin.}\\ \\
\hlruncode{\_, \_, objs, \_ = robot.detect\_objects()}
\hlruncode{compost\_bin\_obj = None}
\hlruncode{recycle\_bin\_obj = None}
\hlruncode{for obj in objs:}
\hlrunlm{\quad is\_match\_description = match\_description(obj, "compost bin", ret\_type="bool")}
\hlruncode{\quad if is\_match\_description:}
\hlruncode{\quad \quad compost\_bin\_obj = obj}
\hlruncode{\quad \quad break}
\hlruncode{for obj in objs:}
\hlrunlm{\quad is\_match\_description = match\_description(obj, "recycle bin", ret\_type="bool")}
\hlruncode{\quad if is\_match\_description:}
\hlruncode{\quad \quad recycle\_bin\_obj = obj}
\hlruncode{\quad \quad break}
\hlruncode{if compost\_bin\_obj is None or recycle\_bin\_obj is None:}
\hlruncode{\quad robot.say("No compost bin or recycle bin found.")}
\hlruncode{else:}
\hlruncode{\quad for obj in objs:}
\hlruncode{\quad \quad if obj == compost\_bin\_obj or obj == recycle\_bin\_obj:}
\hlruncode{\quad \quad \quad continue}
\hlrunlm{\quad \quad is\_compostable = is\_compostable(obj, ret\_type="bool")}
\hlruncode{\quad \quad if is\_compostable:}
\hlruncode{\quad \quad \quad robot.pick\_place(obj, compost\_bin\_obj)}
\hlruncode{\quad \quad else:}
\hlruncode{\quad \quad \quad robot.pick\_place(obj, recycle\_bin\_obj)}
  }}}}}\\
\end{subfigure}

\vspace{2mm}
\hrule
\vspace{2mm}

\begin{subfigure}{.5\linewidth}
\caption{Novel Languages\vspace{-5pt}}
  \setlength{\fboxsep}{2pt}
  \flushright
  \noindent\fbox{\parbox{0.93\linewidth}{\tiny{\texttt{{
  \\
  {Q: Prepare 西红柿炒蛋\ \textcolor{gray}{(stir-fried tomato and eggs)} in the pot.}\\\\
\hlruncode{ingredients = ["西红柿", "鸡蛋"] \textcolor{gray}{["tomato", "egg"]}}
\hlruncode{\_, \_, objs, \_ = robot.detect\_objects()}
\hlruncode{pot\_obj = None}
\hlruncode{for obj in objs:}
\hlrunlm{\quad is\_match\_description = match\_description(obj, "cooking pot", ret\_type="bool")}
\hlruncode{\quad if is\_match\_description:}
\hlruncode{\quad \quad pot\_obj = obj}
\hlruncode{\quad \quad break}
\hlruncode{if pot\_obj is None:}
\hlruncode{\quad robot.say("No cooking pot found.")}
\hlruncode{else:}
\hlruncode{\quad for ingredient in ingredients:}
\hlruncode{\quad \quad for obj in objs:}
\hlrunlm{\quad \quad \quad is\_match\_description = match\_description(obj, ingredient, ret\_type="bool")}
\hlruncode{\quad \quad \quad if is\_match\_description:}
\hlruncode{\quad \quad \quad \quad robot.pick\_place(obj, pot\_obj)}
  }}}}}\\
\end{subfigure}
\hspace{-5pt}
\begin{subfigure}{.5\linewidth}
\caption{Novel Tasks\vspace{-5pt}}
  \setlength{\fboxsep}{2pt}
  \flushright
  \noindent\fbox{\parbox{0.93\linewidth}{\tiny{\texttt{{
  \\
  {Q: My steak is too bland. Can you help?}\\ \\
\hlruncode{robot.say("Sure, let me see what I can do.")}
\hlruncode{\_, \_, objs, \_ = robot.detect\_objects()}
\hlruncode{for obj in objs:}
\hlrunlm{\quad is\_match\_description = match\_description(obj, "steak", ret\_type="bool")}
\hlruncode{\quad if is\_match\_description:}
\hlruncode{\quad \quad steak\_obj = obj}
\hlruncode{\quad \quad break}
\hlruncode{if steak\_obj is None:}
\hlruncode{\quad robot.say("No steak found.")}
\hlruncode{else:}
\hlruncode{\quad robot.say("I'm going to season your steak with some salt and pepper.")}
\hlruncode{\quad robot.pick\_place("salt", steak\_obj)}
\hlruncode{\quad robot.pick\_place("pepper", steak\_obj)}
  }}}}}
\end{subfigure}

\caption{The one-shot prompt as well as the model outputs for a few test instructions for the robotics tasks. When given a single example in the prompt (a), our method can generalize (b-d) to new objects, languages, and task domains. \colorbox{execute-code-red}{Red} highlight indicates LM generated code being executed by the Python interpreter, and \colorbox{execute-lm-purple}{purple} highlight indicates LM simulating the code execution. \textcolor{gray}{Gray text} is for illustration purpose only, and not provided to our model. Note that code in the form of \colorbox{execute-code-red}{robot.<func\_name>} invokes robot APIs.}
\label{fig:robot_exp}
\end{figure}

\begin{figure*}[h]
     \centering
     \begin{subfigure}[b]{0.233\textwidth}
         \centering
         \includegraphics[width=\textwidth]{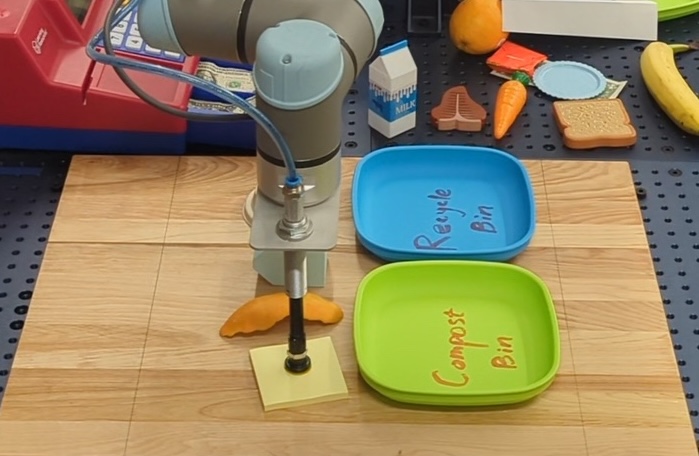}
     \end{subfigure}
     \begin{subfigure}[b]{0.23\textwidth}
         \centering
         \includegraphics[width=\textwidth]{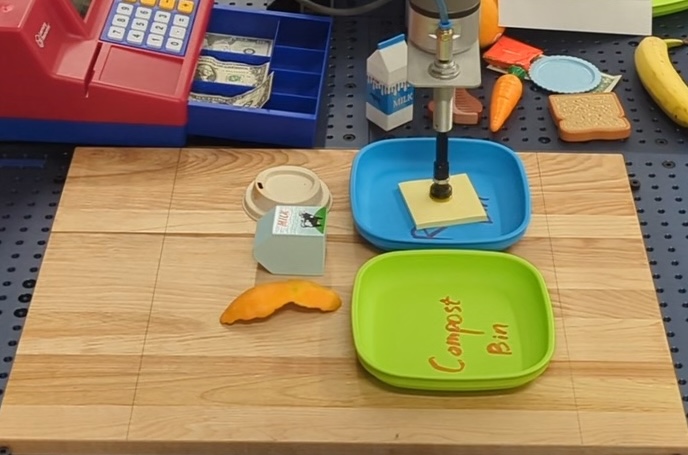}
     \end{subfigure}
     \begin{subfigure}[b]{0.24\textwidth}
         \centering
         \includegraphics[width=\textwidth]{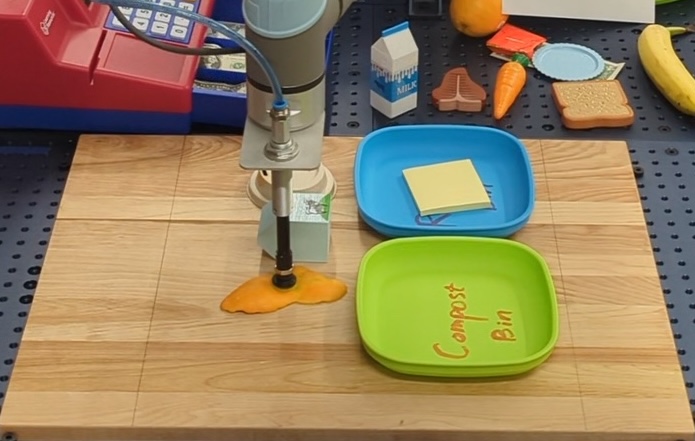}
     \end{subfigure}
     \begin{subfigure}[b]{0.24\textwidth}
         \centering
         \includegraphics[width=\textwidth]{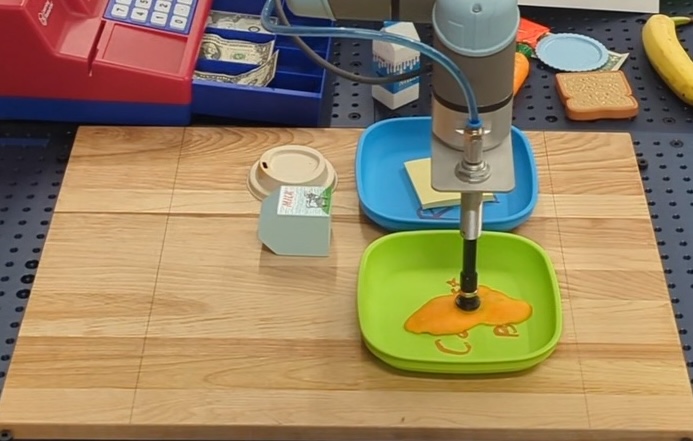}
     \end{subfigure}
     \caption{Robot trajectory visualization for task ``sort the objects on the table into the compost bin and the recycle bin''. \algname first generates code to solve the problem, and then executes the code with Python if possible (e.g., robot APIs like \texttt{detect\_objects} and \texttt{pick\_place}), and with LMulator if not (e.g., commonsense QA like \texttt{is\_compostable}). The robot successfully picks and places the Post-it note to the recycle bin and the orange peel to the compost bin. See the full code in Fig.~\ref{fig:robot_exp}. 
    }
    \label{fig:robot_traj}
\end{figure*}

\begin{figure}[h]
\centering
  \setlength{\fboxsep}{2pt}
  \flushright
  \noindent\fbox{\parbox{0.93\linewidth}{\tiny{\texttt{{
  \\
  {How many countries have I been to? I’ve been to Mumbai, London, Washington, Grand Canyon, Baltimore, Longsheng, Guilin, Beijing, Galapagos, Quito, Barcelona, Paris, Prague, Nice, Dehli, Agra, Rome, Florence, Amalfi, Athens, Míkonos, Málaga, Monaco, Berlin, Munich, Innsbruck, Bern, Milan, Lucerne, Gimmelwald (Schilthornbahn), St Moritz, St Petersburg, Helsinki, Amsterdam, Gdańsk, Vancouver, Anchorage, Montreal, Belize, The Bahamas, Jamaica, Hawaii, Acadia National Park, Stockholm, Copenhagen, Dover, Lyon, Madrid, Toulouse, Santorini, Oslo, Kusadasi, Souda, Rhodes, Tallinn, Venice, Naples, Cape Town, Johannesburg, Addis Abeba, Nairobi, Seattle, San Francisco, Chicago, St Louis, Memphis, Chinle, Stanford, New York, Philadelphia, Boston, Miami, New Orleans, Walt Disney World Resort, Jacksonville, Las Vegas, Los Angeles, Portland, Salt Lake City, Tahoe City, Phoenix, Albuquerque, Cleveland, Charlottesville, Nags Head, Newfoundland and Labrador, Burlington, Wilmington, Myrtle Beach, St Lucia, Barbados, Banff, Haiti, Montego Bay, Sao Palo, Rio, Lima, Cusco, Cozumel, Amarillo, Yosemite National Park, Joshua Tree, Zion National Park, Bryce Canyon National Park, Grand Teton National Park, Yellowstone National Park, Glacier National Park, Mount Hood, Paso Robles, San Diego, Bend, North Cascades National Park, Olympic National Park Visitor Center, Jasper National Park, Sequoia National Park, Kings Canyon National Park, Shasta National Forest, Mount Saint Helens, Mount Rainier, Austin, Buenos Aires, El Calafate, El Chaltén, Fitz Roy, Torres del Paine National Park, Puerto Natales, Puerto Varas, Santiago, Marble Caves, Cerro Castillo, Coyhaique, Singapore, Casablanca, Marrakesh, Cairo, Jerusalem, Tokyo, Kyoto Prefecture, Taipei City, Taichung City, Krk, Naturpark Puez-Geisler, Ljubljana, Plitvice Lakes National Park, Fairbanks, Juneau, Dallas, Sydney, Cairns, Brisbane, Hook Island, Charleston, Panama City, Bangkok, Chiang Mai, Bengaluru, Denver, Indianapolis, Nashville, Blacksburg, Lisbon, Porto, Estes Park, Coeur d'Alene, Hood River, Denali, Sitka, Mexico City, Warsaw, Geneva, Auckland, Queenstown, Whitefish, Minneapolis, Sioux Falls, Bozeman, Missoula, Springfield, Skye, Edinburgh, Honolulu, Kauai, Haleakalā National Park, Wrangell-St. Elias National Park \& Preserve, Atlanta, Tirana, Corfu, Siena.}}}}}}
  \caption{Full question used in Fig.~\ref{fig:intro}}
\label{fig:intro_query}
\end{figure}



\end{CJK*}
\end{document}